\documentclass[10pt,twocolumn,letterpaper]{article}

\usepackage[pagenumbers]{cvpr}

\usepackage{graphicx}
\usepackage{amsmath}
\usepackage{amssymb}
\usepackage{booktabs}

\usepackage{times}
\usepackage{epsfig}
\usepackage{textcomp}
\usepackage{overpic}
\usepackage{makecell}
\usepackage{bbding}

\usepackage{soul}

\usepackage{times}
\usepackage{epsfig}
\usepackage{float}
\usepackage{placeins}
\usepackage{booktabs}
\usepackage{color, colortbl}
\usepackage{stfloats}
\usepackage{enumitem}
\usepackage{tabularx}
\usepackage{xstring}
\usepackage{multirow}
\usepackage{xspace}
\usepackage{url}
\usepackage{xcolor}
\usepackage[hang,flushmargin]{footmisc}
\usepackage[accsupp]{axessibility}
\usepackage{hyperref}

% ...

\graphicspath{{./figs/}}

\newcommand{\R}[1]{{%
    \textbf{%
        \ifstrequal{#1}{1}{\textcolor{red}{R#1}}{%
        \ifstrequal{#1}{2}{\textcolor{blue}{R#1}}{%
        \ifstrequal{#1}{3}{\textcolor{magenta}{R#1}}{%
        \ifstrequal{#1}{4}{\textcolor{teal}{R#1}}{%
                           \textcolor{cyan}{R#1}%
        }}}}%
    }%
}}

\usepackage{xr-hyper}

\makeatletter
\newcommand*{\addFileDependency}[1]{
  \typeout{(#1)}
  \@addtofilelist{#1}
  \IfFileExists{#1}{}{\typeout{No file #1.}}
}

\makeatother

\usepackage[accsupp]{axessibility}  
\usepackage[capitalize]{cleveref}
\crefname{section}{Sec.}{Secs.}
\crefname{table}{Table}{Tables}
\crefname{figure}{Fig.}{Figs.}

\begin{document}

\title{Multi-view Inverse Rendering for Large-scale Real-world Indoor Scenes}
\author{
    {Zhen Li}\textsuperscript{1}  $\qquad$
    {Lingli Wang}\textsuperscript{1}  $\qquad$
    {Mofang Cheng}\textsuperscript{1}  $\qquad$
    {Cihui Pan}\textsuperscript{1,}\footnotemark[1]  $\qquad$
    {Jiaqi Yang}\textsuperscript{2,}\thanks{Co-corresponding authors.}  \\
    \textsuperscript{1}Realsee $\qquad$ \textsuperscript{2}Northwestern Polytechnical University\\
    {\tt\small yodlee@mail.nwpu.edu.cn,\{wanglingli008,chengmofang001,pancihui001\}@realsee.com,jqyang@nwpu.edu.cn}
}
\maketitle

\begin{abstract} We present a efficient multi-view inverse rendering method for large-scale real-world indoor scenes that reconstructs global illumination and physically-reasonable SVBRDFs. Unlike previous representations, where the global illumination of large scenes is simplified as multiple environment maps, we propose a compact representation called \textbf{T}exture-\textbf{b}ased \textbf{L}ighting (TBL). It consists of 3D mesh and HDR textures, and efficiently models direct and infinite-bounce indirect lighting of the entire large scene. Based on TBL, we further propose a hybrid lighting representation with precomputed irradiance, which significantly improves the efficiency and alleviates the rendering noise in the material optimization. To physically disentangle the ambiguity between materials, we propose a three-stage material optimization strategy based on the priors of semantic segmentation and room segmentation. Extensive experiments show that the proposed method outperforms the state-of-the-art quantitatively and qualitatively, and enables physically-reasonable mixed-reality applications such as material editing, editable novel view synthesis and relighting.
The project page is at \href{https://lzleejean.github.io/TexIR}{https://lzleejean.github.io/TexIR.} \end{abstract}
\section{Introduction}
\label{sec:intro}

Inverse rendering aims to reconstruct geometry, material and illumination of an object or a scene from images. These properties are essential to downstream applications such as scene editing, editable novel view synthesis and relighting. 
However, decomposing such properties from the images is extremely ill-posed, because different configurations of such properties often lead to similar appearance.
With recent advances in differentiable rendering and implicit neural representation, several approaches have achieved significant success on small-scale object-centric scenes with explicit or implicit priors~\cite{nerv2021,nerfactor,boss2021nerd,Luan2021UnifiedSA,iron-2022,zhang2022invrender,munkberg2021nvdiffrec,yao2022neilf,yang2022psnerf}. However, inverse rendering of large-scale indoor scenes has not been well solved. 


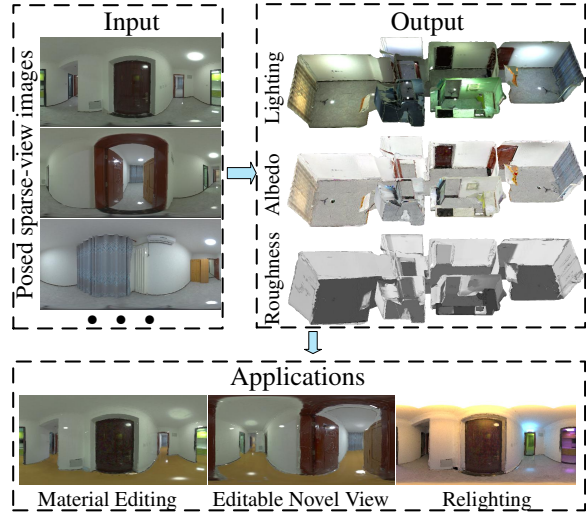
\begin{figure}[t]
    \centering
    
    
    
    
    \begin{overpic}[scale=0.22]{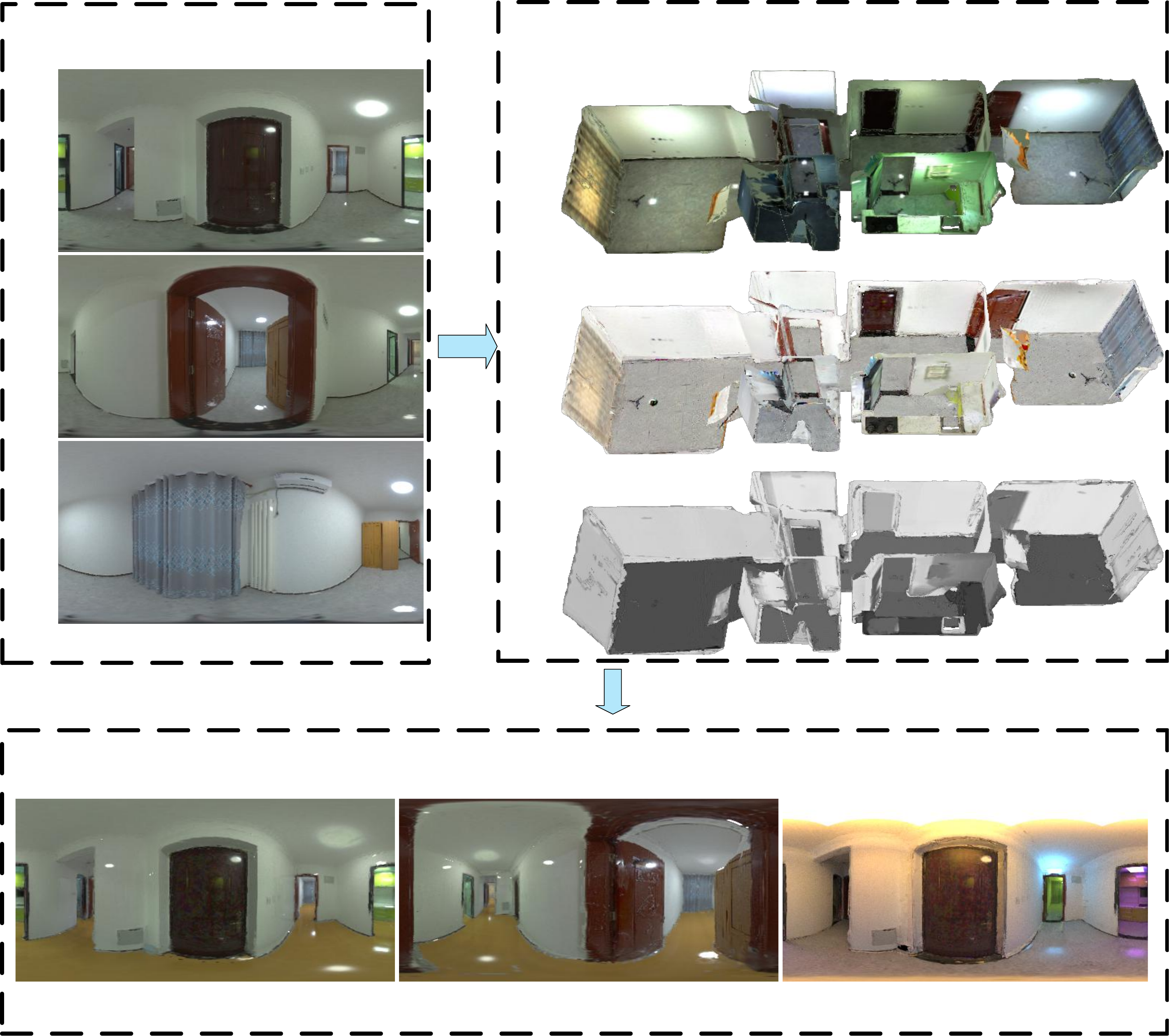}
    \put(0.7,37){\rotatebox{90}{\small Posed sparse-view images}}
    \put(44,32.5){\rotatebox{90}{\footnotesize Roughness}}
    \put(44,51.5){\rotatebox{90}{\footnotesize Albedo}}
    \put(44,67.5){\rotatebox{90}{\footnotesize Lighting}}
    
    \put(13,32.5){{\footnotesize $\bullet$}}
    \put(18,32.5){{\footnotesize $\bullet$}}
    \put(23,32.5){{\footnotesize $\bullet$}}
    
    \put(4.5,1){\footnotesize Material Editing}
    \put(35,1){\footnotesize Editable Novel View}
    \put(75,1){\footnotesize Relighting}
    
    \put(16,84){\normalsize Input}
    \put(66,84){\normalsize Output}
    \put(38,22){\normalsize Applications}

    \end{overpic}
    \caption{Given a set of posed sparse-view images for a large-scale scene, we reconstruct global illumination and SVBRDFs. The recovered properties are able to produce convincing results for several mixed-reality applications such as material editing, editable novel view synthesis and relighting. Note that we change roughness of all walls, and albedo of all floors. The detailed specular reflectance shows that our method successfully decomposes physically-reasonable SVBRDFs and lighting. Please refer to supplementary videos for more animations.}
    \label{fig:fig1}
\end{figure}

There are two main challenges for large-scale indoor scenes. 1) \textbf{\textit{Modelling the physically-correct global illumination.}}
There are far more complex lighting effects, \eg, inter-reflection and cast shadows, in large-scale indoor scenes than object-centric scenes due to complex occlusions, materials and local light sources. Although the widely-used image-based lighting (IBL) is able to efficiently model direct and indirect illumination, it only represents the lighting of a certain position~\cite{debevec2008rendering, neuralSengupta19, gardner-sigasia-17, Garon_2019_CVPR}. The spatial consistency of per-pixel or per-voxel IBL representations~\cite{Zhou_2019_ICCV, li2020inverse, li2022phyir,wang2021learning} is difficult to ensure. Moreover, such incompact representations require large memory.
Parameterized lights\cite{gardner2019deep, li2022editing} such as point light, area light and directional light are naturally globally-consistent, but modeling the expensive global light transport will be inevitable \cite{azinovic2019inverse, nimierdavid2021material, zhang2022invrender}.
Thus, simple lighting representations applied in previous works are unsuitable in large-scale scenes. 
2) \textbf{\textit{Disentangling the ambiguity between materials.}} Different configurations of materials often lead to similar appearance, and to add insult to injury, there are an abundance of objects with complex and diverse materials in large-scale scenes. In object-centric scenes, dense views distributed on the hemisphere are helpful for alleviating the ambiguity~\cite{MobileSVBRDF2018,multiSVBRDF2019,jointSVBRDF2019,physg2020,Luan2021UnifiedSA,iron-2022}. However, only sparse views are available in large-scale scenes, which more easily lead to ambiguous predictions~\cite{nimierdavid2021material, yao2022neilf}.

In this work, we present TexIR, an efficient inverse rendering method for large-scale indoor scenes.
Aforementioned challenges are tackled individually in the following.
1) We model the infinite-bounce global illumination of the entire scene with a novel compact lighting representation, called TBL. The TBL is able to efficiently represent the infinite-bounce global illumination of any position within the large scene. Such a compact and explicit representation provides more physically-accurate and spatially-varying illumination to guide the material estimation. 
Directly optimizing materials with TBL leads to expensive computation costs caused by high samples of the monte carlo sampling. Therefore, 
we precompute the irradiance based on our TBL, which significantly accelerates the expensive computation in the material optimization process.
2) To ameliorate the ambiguity between materials, 
we introduce a segmentation-based three-stage material optimization strategy. 
Specifically, we optimize a coarse albedo based on Lambertian-assumption in the first stage. In the second stage, we integrate semantics priors to guide the propagation of physically-correct roughness in regions with same semantics. In the last stage, we fine-tune both albedo and roughness based on the priors of semantic segmentation and room segmentation.
By leveraging such priors, physically-reasonable albedo and roughness are disentangled globally.

To summarize, the main contributions of our method are as follows:
\begin{enumerate}
    \item A compact lighting representation for large-scale scenes, where the infinite-bounce global illumination of the entire large scene can be handled efficiently.
    \item A segmentation-based material optimization strategy to globally and physically disentangle the ambiguity between albedo and roughness of the entire scene.
    \item A hybrid lighting representation based on the proposed TBL and precomputed irradiance to improve the efficiency in the material optimization process.
\end{enumerate}


\begin{figure*}[htbp]
    \centering
    
    
    
    
    \begin{overpic}[scale=0.19]{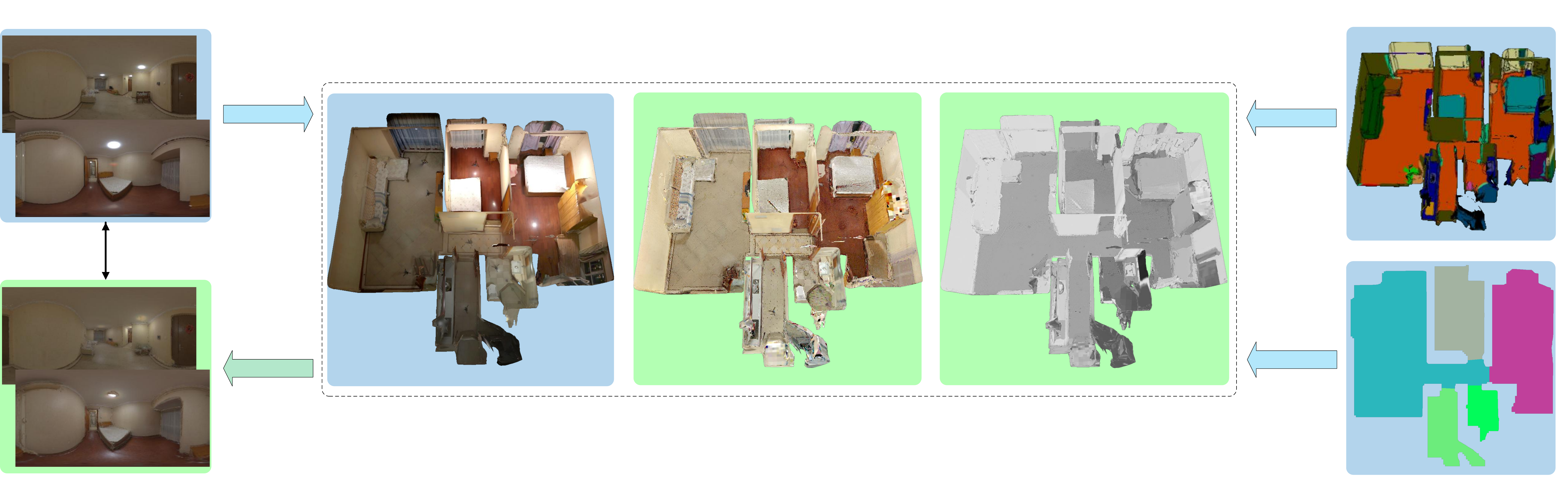}
    
    \put(-1,30){{\small Sparse posed images}}
    \put(26,27){{Lighting}}
    \put(46.5,27){{Albedo}}
    \put(64.5,27){{Roughness}}
    
    \put(-1,-1){{\small Re-rendered images}}
    
    \put(15,25){{\small MVS}}
    \put(15,21){{\small HDR}}
    \put(14,19.5){{\small Texture}}
    \put(16,9){{\small DR}}
    
    \put(79,25){\footnotesize semantics}
    \put(81,21){\footnotesize prior}
    \put(80.8,9.5){\footnotesize room}
    \put(81,5.5){\footnotesize prior}
    
    
    \put(3,14.5){{\footnotesize Loss}}

    \end{overpic}
    
    \caption{\textbf{Overview of our inverse rendering pipeline.} Given sparse calibrated HDR images for a large-scale scene, we reconstruct the geometry and HDR textures as our lighting representation. PBR material textures of the scene, including albedo and roughness, are optimized by differentiable rendering (DR). The ambiguity between materials is disentangled by the semantics prior and the room segmentation prior. Gradient flows in \textcolor[RGB]{0,200,0}{Green Background}.}
    \label{fig:framework}
\end{figure*}
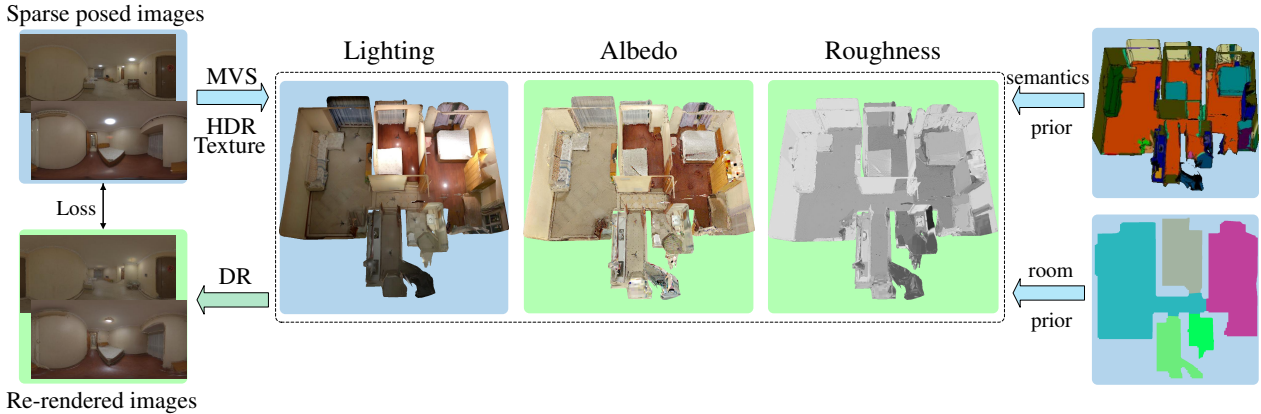

\section{Related Work}
\label{sec:related}

\noindent \textbf{Image-based lighting.}
 Debevec \etal~\cite{debevec2008rendering} first introduced the high dynamic range (HDR) light probe as a omni-directional lighting representation of a certain position, called IBL, which plays a vital role on inverse rendering and lighting estimation tasks in the computer vision. Barron and Malik \cite{SIRFS} fitted the spherical-harmonic (SH) illumination, a parameterized representation of IBL, from a single image. Then they extended this single SH illumination into a set of SH illumination \cite{barron2013intrinsic}. Zhou \etal \cite{Zhou_2019_ICCV} predicted the per-pixel SH lighting from a single image. Further, the per-pixel spherical-gaussian (SG) lighting \cite{li2020inverse} and the per-pixel light probe \cite{li2022phyir} are applied. Wang \etal \cite{wang2021learning} proposed a per-voxel SG lighting in the 3D indoor scene. The spatial consistency of illumination is unable to ensure with above lighting representations, especially for 3D large-scale scenes. Although light probes are able to project consistent light probes in others positions with known geometry, the efficiency in the optimization process is limited \cite{li2021lighting}. Our proposed TBL not only reserves the advantage of IBL, modelling infinite-bounce global illumination efficiently, but also is naturally globally-consistent.

\noindent \textbf{Parametric lighting.}
Parametric lights, such as point lights, spot lights, area lights and directional lights, are classical lighting representations to define the illumination of a scene in computer graphics. 
Most methods only use one of the above lighting to represent the illumination of a scene. Nestmeyer \etal~\cite{nestmeyer2020faceRelighting} predicted a directional light from a face image. Junxuan Li and Hongdong Li \cite{li2022shape} used different directional lights as their illumination setting. Several approaches \cite{bi2020nrf, nerv2021, Luan2021UnifiedSA, iron-2022} model the illumination of a static object with point lights. Li \etal~\cite{li2022editing} predicted area lights and SG directional lights for a single indoor image. Zhang \etal~\cite{zhang2016emptying} optimized a more complex configuration via vertex-based lighting, including point lights, line lights, area lights and a environment map, but this method require user inputs to label where is the light source. Our TBL leverages the HDR texture of the entire scene to represent the illumination at any position within this 3D scene without any user input. Moreover, the TBL is compact and naturally globally-consistent.

\noindent \textbf{Implicit lighting.}
With great advances in implicit representation, researchers began to use implicit neural networks to represent scenes. In particular, NeRF \cite{mildenhall2020nerf} has shown impressive results on scene representation. It leverages a neural density field and a neural radiance field to model a static small-scale scene. Zhang \etal \cite{zhang2022invrender} obtained the incident radiance of a certain position from a pre-trained outgoing radiance field\cite{yariv2020multiview}. To go a step further, the neural incident radiance field (NIRF), which represents incident radiance from any direction at any position, have shown great potential for inverse rendering \cite{yao2022neilf} and novel view synthesis \cite{wang2022r2l}. 
However, without constraints for such powerful implicit representation, the ambiguity between material and lighting is difficult to disentangle. Our explicit TBL is capable of eliminating this ambiguity in a interpretable manner.

\noindent \textbf{Material estimation.}
The deep neural network has achieved significant success in many vision tasks with large-scale real-world datasets, such as object detection, semantic segmentation and depth estimation. Unfortunately, collecting the annotation of materials in real-world is difficult at scale. IIW \cite{iiwdataset} only labels sparse pairwise reflectance comparison for a single image. Therefore, most methods~\cite{li2018cgintrinsics,li2018learning,li2020inverse,neuralSengupta19,TwoShotShapeAndBrdf,wang2021learning,li2022phyir,irisformer2022} are alternative to use synthetic dataset to disambiguate the ambiguous properties.
These approaches training on synthetic datasets struggle to eliminate the inevitable domain gap though several datasets have achieved photo-realistic results~\cite{openrooms, li2022phyir, roberts2021hypersim}. Optimization-based methods \cite{SIRFS,barron2013intrinsic,physg2020,nerv2021,boss2021nerd,munkberg2021nvdiffrec,zhang2022invrender,nimierdavid2021material} have shown impressive results on real-world multi-view images. It is known that the prior of different materials is essential to be leveraged in the optimization process to alleviate the ambiguity between materials. Barron and Malik \cite{SIRFS} applied a smoothness term for albedo. Multiple variants of filters are used to smooth material in follow-up works \cite{bisai18intrinsic,li2020inverse,li2022phyir}. Zhang \etal~\cite{nerfactor} leveraged learning priors from real-world BRDFs and imposed a spatial smoothness prior in the latent space. A similar idea has been adopted by a recent work \cite{zhang2022invrender}. Schmitt \etal~\cite{Schmitt_2020_CVPR} and Luan \etal~\cite{Luan2021UnifiedSA} assumed similar diffuse albedos to have similar specular ones, and applied a non-local bilateral regularizer for specular albedo. Compared to object-centric scenes \cite{nerv2021,nerfactor,boss2021nerd,physg2020,iron-2022,zhang2022invrender,munkberg2021nvdiffrec}, disentangling the ambiguity between materials is more challenging in the large-scale scene due to sparser observations and wider variety of objects and reflectance properties. Zhang \etal~\cite{zhang2016emptying} recovered a constant attribute for each floor, wall and ceiling of an empty room. Most similar to ours, Nimier-David \etal~\cite{nimierdavid2021material} and Haefner \etal~\cite{Haefner213dv} also assumed similar semantic regions to have similar roughness and specularity. The first difference is we optimize the roughness in a soft manner instead optimizing a constant value \cite{nimierdavid2021material, Haefner213dv}. The another difference is we leverage the room segmentation prior, which enables us to recover different roughness even for similar semantic regions. Thanks to our efficient global illumination representation and explicit optimization strategy, our method is 20$\times$ faster than the differentiable path tracing-based method~\cite{nimierdavid2021material}.



\section{Methodology}
As shown in Fig.~\ref{fig:framework}, given a set of calibrated HDR images of a large-scale indoor scene, our method aims to accurately recover globally-consistent illumination and Spatially-Varying Bidirectional Reflectance Distribution Functions (SVBRDFs), which can be conveniently integrated into graphics pipelines and downstream applications.  To this end, we propose TBL to represent the global illumination of large-scale indoor scenes (Sec.~\ref{sec:tbl}). In order to improve the optimizing efficiency and quality of re-rendered images in the material estimation stage, we adopt a hybrid lighting representation based on our TBL (Sec.~\ref{sec:hybrid_lighting}). To reconstruct physically-reasonable SVBRDFs, we present a segmentation-based three-stage material estimation strategy (Sec.~\ref{sec:material}), which can handle ambiguity of materials in complex large-scale indoor scenes very well.


\subsection{Texture-based Lighting}
\label{sec:tbl}
We address the issue of how to represent the illumination of a large-scale indoor scene with TBL. The advantages of TBL respectively are the compactness of neural representation, the global illumination of IBL, and the interpretability and spatial consistency of parametric lights.

The proposed TBL, which is a global representation for the entire scene, defines the outgoing radiance for all surface points. 
We assume that only the diffuse lighting exists in the scene since the diffuse lighting often dominates the scene, similar to radiosity-based methods~\cite{yu99sig, wood00slf, di88}. Therefore, the outgoing radiance of one point is typically equal to the value of the HDR texture, \ie, the observed HDR radiance of corresponding pixels in input HDR images. 

We initially reconstruct the mesh model of a entire large-scale scene with off-the-shelf classical MVS techniques, \eg, colmap \cite{colmap}.
Finally, we reconstruct the HDR texture based on the input HDR images. Therefore, the global illumination is queried from any direction at any position through the HDR texture, as shown in Fig.~\ref{fig:tbl_incident}. 

\subsection{Hybrid Lighting Representation}
\label{sec:hybrid_lighting}
According to our proposed TBL, the render equation~\cite{kajiya1986rendering} can be rewritten as follow:
\begin{equation}\label{eq:re}
L_{o}(x, \omega_o) = \int_{H^{+}} f_r(x, \omega_i, \omega_o) Q(x, \omega_i, G, T_{hdr}) (\omega_i \cdot n) d\omega_i
\end{equation}
where $H^{+}$ denotes hemisphere; $x$ denotes a surface point; $\omega_i$ denotes inverse incident light direction; $\omega_o$ denotes view direction; $n$ denotes normal; $f_r$ denotes the BRDF of point $x$; $Q$ denotes the HDR lighting of the intersection point between the known geometry $G$ and the ray $r(t) = x + t\omega_i$, queried by HDR textures $T_{hdr}$.

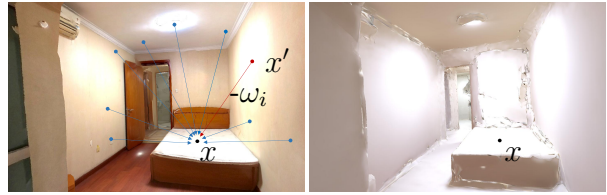
\begin{figure}[t]
    \centering
    \begin{overpic}[scale=0.24]{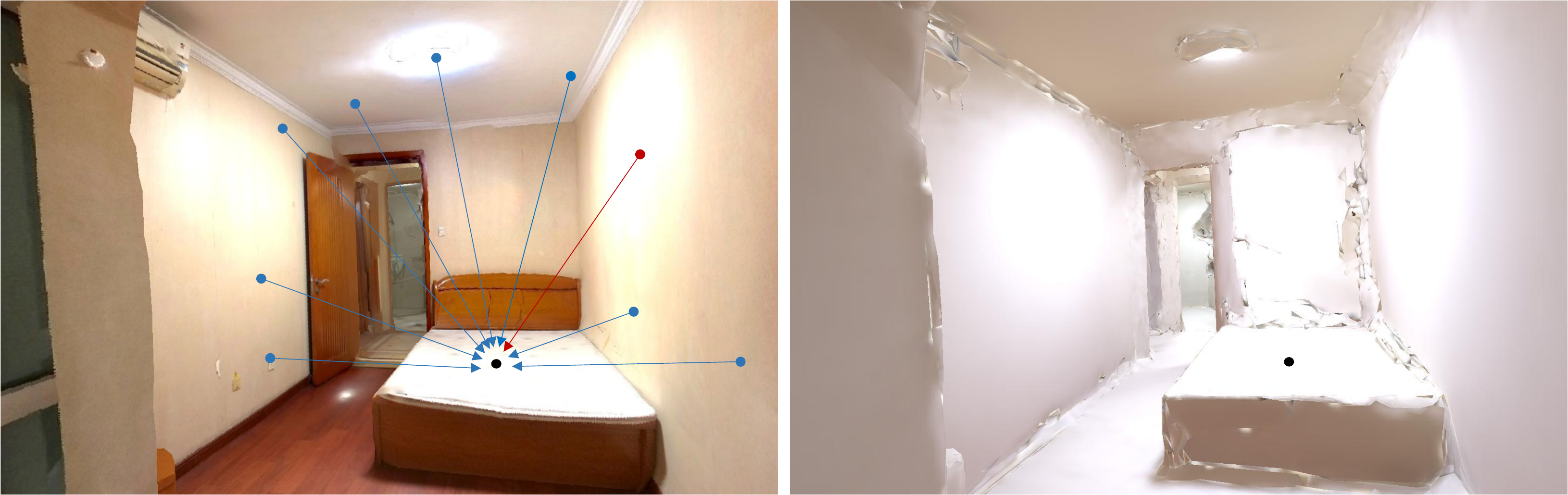}
    \put(32,5){\large $x$}
    \put(37,15){\large -$\omega_i$}
    \put(43,20){\large $x'$}
    
    \put(83,5){\large $x$}
    
    \end{overpic}
    \caption{\textbf{Visualization of TBL (left) and precomputed irradiance (right).} For any surface point $x$, the incident radiance from direction $-\omega_i$ can be queried from the HDR texture of the point $x'$, which is the intersection point between the geometry and the ray $r(t)=x+t\omega_{i}$. The irradiance can be directly queried from the precomputed irradiance of $x$ via NIrF or IrT.}
    \label{fig:tbl_incident}
\end{figure}

In practice, we calculate the Monte Carlo numerical integration with importance sampling~\cite{realshading}. To decrease the variance, a large sample number seems to be inevitable, which will significantly increase computation cost and  memory cost in the optimization process. 
Inspired by precomputed radiance transfer~\cite{Greger1998TheIV}, we precompute irradiance of surface points for the diffuse component. 
Therefore, the irradiance is queried efficiently without expensive online computation as shown in Fig.~\ref{fig:tbl_incident} (right)~\cite{di88}.
The Eq.\ref{eq:re} can be rewritten as:
\begin{equation}\label{eq:re_d+s}
L_{o}(x, \omega_o) = L_{d}(x, \omega_o) + L_{s}(x, \omega_o)
\end{equation}
where $L_{d}$ denotes the diffuse component and $L_{s}$ denotes the specular component. The $f_r$ of $L_{s}$ would be modified as $f_s$. The detailed formulation can be found in the supplementary material.

We propose two representations to model precomputed irradiance. One is a \textbf{n}eural \textbf{ir}radiance \textbf{f}ield (NIrF), a shallow Multi-Layer-Perceptrons (MLP). It takes a surface point $p$ as input and outputs the irrdiance of $p$. Another one is a \textbf{ir}radiance \textbf{t}exture (IrT), similar to the light map widely used in computer graphics. Based on such hybrid lighting representation consists of precomputed irradiance for the diffuse component and source TBL for the specular component, the rendering noise significantly decrease and the materials are optimized efficiently.
Therefore, the diffuse component of Eq.~\ref{eq:re_d+s} can be modeled as Eq.~\ref{eq:ir}.
\begin{equation}\label{eq:ir}
L_{d}(x, \omega_o) = f_d(x) Ir(x)
\end{equation}
where $f_d$ denotes the diffuse BRDF of point $x$ and $Ir$ denotes the irradiance of point $x$.

\begin{table*}[t]
	\centering \caption{\textbf{Quantitative comparison on our synthetic dataset.} Our method significantly outperforms the state-of-the-arts in roughness estimation. NeILF$*$~\cite{yao2022neilf} denotes source method with their implicit lighting representation.}\label{tb:syn}
    \scalebox{0.85}{
    \begin{tabular}{*{13}{c}}
      \toprule
      \multirow{2}*{Method} & \multicolumn{3}{c}{Albedo} & \multicolumn{3}{c}{Roughness} & \multicolumn{3}{c}{Novel view synthesis} & \multicolumn{3}{c}{Re-rendering}\\
      \cmidrule(lr){2-4}\cmidrule(lr){5-7}\cmidrule(lr){8-10}\cmidrule(lr){11-13}
      & PSNR$\uparrow$ & SSIM$\uparrow$ & MSE$\downarrow$ & PSNR$\uparrow$ & SSIM$\uparrow$ & MSE$\downarrow$ & PSNR$\uparrow$ & SSIM$\uparrow$ & MSE$\downarrow$ & PSNR$\uparrow$ & SSIM$\uparrow$ & MSE$\downarrow$ \\
      \midrule
      PhyIR~\cite{li2022phyir} &11.9726&0.6880&0.0635 &12.5468&0.7671&0.0556 &-&-&- &-&-&- \\
      InvRender~\cite{zhang2022invrender} &16.9760&0.6305&0.0201 &9.1806&0.4787&0.1208 &22.2771&0.7826&0.0059 &24.2851&0.7834&0.0037 \\
      NVDIFFREC~\cite{munkberg2021nvdiffrec} &\textbf{21.2551}&0.8100&\textbf{0.0075} &7.6269&0.1348&0.1727 &23.4959&0.9019&0.0045 &29.7279&0.9323&0.0011 \\
      NeILF$*$~\cite{yao2022neilf} &14.2137&0.5184&0.0379 &11.5778&0.5974&0.0695 &22.3765&0.7598&0.0058 &25.1092&0.7654&0.0031 \\
      NeILF~\cite{yao2022neilf} &17.0707&0.6489&0.0196 &11.1654&0.7099&0.0765 &22.0703&0.7823&0.0062 &24.4710&0.7857&0.0036 \\
      Ours  &20.4169&\textbf{0.8514}&0.0091 &\textbf{20.2132}&\textbf{0.9161}&\textbf{0.0095} &\textbf{25.0462}&\textbf{0.9264}&\textbf{0.0031}  &\textbf{34.2669}&\textbf{0.9635}&\textbf{0.0004}  \\
      \bottomrule
    \end{tabular}
    }
\end{table*}

\subsection{Segmentation-based Material Estimation}
\label{sec:material}
Instead of optimizing neural material~\cite{boss2021nerd, nerv2021,nerfactor,physg2020,zhang2022invrender,yao2022neilf,yang2022psnerf}, which is hard to model a large-scale scene with extremely complex materials and is mismatched to the traditional graphics engines,  we directly optimize explicit material textures of the geometry.

We use the simplified Disney BRDF model~\cite{Burley2012pbrdisney} with Spatially-Varying (SV) albedo 
and SV roughness 
as parameters. Optimizing explicit material textures straightforwardly leads to inconsistent and unconverged roughness due to sparse observations~\cite{zhang2022invrender}. We address this problem by leveraging the priors of semantics and room segmentation. 
The semantic images are predicted by learning-based models~\cite{mmseg2020} and room segmentation is calculated by the occupancy grid~\cite{Elfes1989UsingOG}.
Our segmentation-based strategy has three phases. Details of each stage are described in the following subsections.

\noindent \textbf{Stage \uppercase\expandafter{\romannumeral1}: albedo initialization.}
We optimize a coarse albedo based on Lambertian assumption instead of initializing the albedo as a constant, which is widely used in object-centric scenes~\cite{Luan2021UnifiedSA}. According to estimated illumination in Sec.~\ref{sec:hybrid_lighting}, we can directly calculate the albedo through Eq.~\ref{eq:ir}. However, it recovers over-bright albedo on the highlight regions, which leads to high roughness in the next stage. Therefore, we apply a semantic smoothness constraint to encourage that the albedo/roughness/feature is as close to the mean within the class as possible:
\begin{equation}\label{eq:semantic smoothness}
\mathcal{L}_{ss} = \sum_{c} \bigg\vert F - \frac{\sum_{p} F\odot M_{seg}(c)}{\sum_{p}M_{seg}(c) + \epsilon}  \bigg\vert \odot M_{seg}(c)
\end{equation}
where $L_{ss}$ denotes the semantic smoothness loss; $c$ denotes one of the classes of semantics; $F$ denotes the feature image to be smoothed and we use the image-space diffuse albedo $A$ in this stage; $p$ denotes the pixel of $F$; $\odot$ is an element-wise product; $M_{seg}$ denotes the mask of semantic segmentation;  $\epsilon$ denotes a tiny number. The coarse albedo is optimized by:
\begin{equation}\label{eq:stage1}
\mathcal{L}_{albedo} = \vert I - L_{d} \vert + \beta_{ssa}\mathcal{L}_{ss}
\end{equation}
where $\beta_{ssa}$ denotes the weight of semantic smoothness loss for albedo, and $I$ denotes the input HDR images.

\noindent \textbf{Stage \uppercase\expandafter{\romannumeral2}: VHL-based sampling and semantics-based propagation.}
In multi-view images, only sparse specular cues are observed, which lead to globally inconsistent roughness~\cite{zhang2022invrender}, especially for large-scale scenes. As shown in Baseline of Fig.~\ref{fig:ablation_material_main} and NVDIFFREC\cite{munkberg2021nvdiffrec} of Fig.~\ref{fig:comparison_real2_main}, only the roughness of highlight regions is optimized reasonably. Therefore, by leveraging the prior of semantic segmentation, the reasonable roughness of highlight regions would be propagated into the regions with same semantics.

We first render images based on input poses with roughness 0.01 to find \textbf{v}irtual \textbf{h}igh\textbf{l}ight (VHL) regions for each semantic class. Then, we optimize the roughness on these VHL regions according to the \textit{frozen} coarse albedo and illumination. Meanwhile, the reasonable roughness can be propagated into the same semantic segmentation through Eq.\ref{eq:propagation}:
\begin{small}
\begin{equation}\label{eq:propagation}
\mathcal{L}_{sp} = \sum_{c} \bigg\vert R - quantile(R \odot M_{vhl}(c),q) \bigg\vert \odot (M_{seg}(c)-M_{vhl}(c))
\end{equation}
\end{small}
where $L_{sp}$ denotes the semantics-based propagation loss; $R$ denotes the image-space roughness; $quantile$ denotes the $q$-th quantiles on VHL regions and we set the $q$ as 0.4 for robustness; $M_{vhl}$ denotes the mask of VHL regions for each class. The roughness can be optimized by:
\begin{equation}\label{eq:stage2}
\mathcal{L}_{roughness} = \vert I - L_{o} \vert + \beta_{sp}\mathcal{L}_{sp}
\end{equation}

\noindent \textbf{Stage \uppercase\expandafter{\romannumeral3}: Segmentation-based fine-tuning.}
In the last phase, we fine-tune all material textures based on the priors of semantic segmentation and room segmentation. Specifically, we apply a similar smoothness constraint to Eq.~\ref{eq:semantic smoothness} and a room smoothness constraint for roughness, which makes the roughness of different rooms smoother in a soft manner. The room smoothness constraint is formulated by Eq.~\ref{eq:room smoothness}:

\begin{equation}\label{eq:room smoothness}
\mathcal{L}_{rs} = \sum_{c} \bigg\vert R - \frac{\sum_{p} R\odot M_{room}(c)}{\sum_{p}M_{room}(c) + \epsilon}  \bigg\vert \odot M_{room}(c)
\end{equation}
where $L_{rs}$ denotes the room segmentation-based smoothness loss; $M_{room}$ denotes the mask of different room segmentation; $c$ denotes one of the index of each room.

We do not apply any smoothness constraint for albedo. The total loss is defined as:
\begin{equation}\label{eq:stage3}
\mathcal{L}_{all} = \vert I - L_{o} \vert + \beta_{ssr}(\mathcal{L}_{ss}+\mathcal{L}_{rs})
\end{equation}
where $\beta_{ssr}$ denotes the weight of segmentation smoothness loss for roughness and we use the image-space roughness $R$ in $\mathcal{L}_{ss}$ in Stage \uppercase\expandafter{\romannumeral3}.

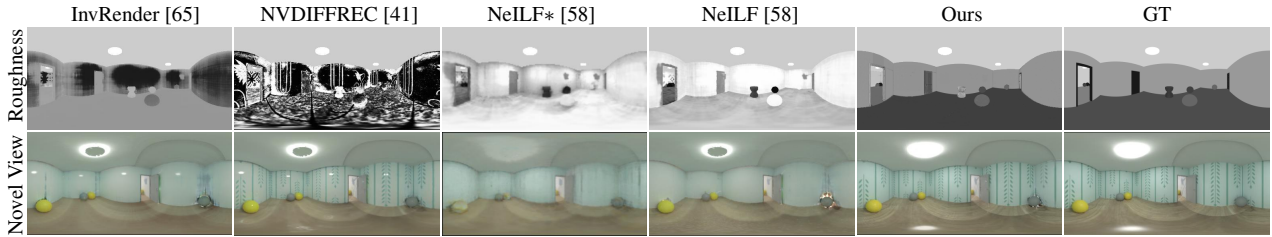
\begin{figure*}[t]
    \centering
    \begin{overpic}[scale=0.251]{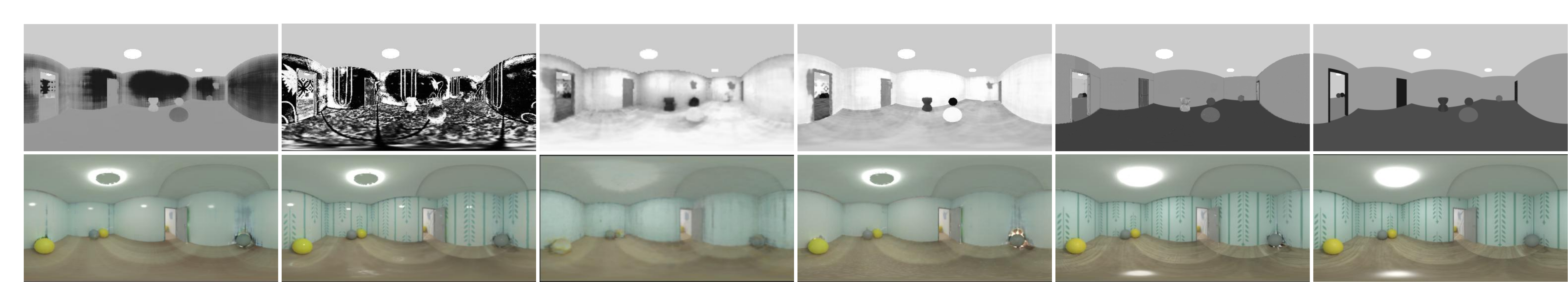}
    \put(5,17){\footnotesize InvRender\cite{zhang2022invrender}}
    \put(20,17){\footnotesize NVDIFFREC\cite{munkberg2021nvdiffrec}}
    \put(38,17){\footnotesize NeILF$*$\cite{yao2022neilf}}
    \put(55,17){\footnotesize NeILF\cite{yao2022neilf}}
    \put(74,17){\footnotesize Ours}
    \put(90,17){\footnotesize GT}
    
    \put(0,9){\rotatebox{90} {\footnotesize Roughness}}
    \put(0,0){\rotatebox{90} {\footnotesize Novel View}}
    
    \end{overpic}
    \caption{\textbf{Qualitative comparison on synthetic dataset.} Our method is able to produce realistic specular reflectance. NeILF$*$~\cite{yao2022neilf} denotes source method with their implicit lighting representation.}
    \label{fig:syn_roughness}
\end{figure*}

\section{Experiments}


\subsection{Datasets}
\label{sec:exp_dataset}

\noindent \textbf{Synthetic dataset.} We create a synthetic scene with diverse material and light sources with a path tracer\cite{li2020inverse}. We render 24 views for optimization and 14 views as novel views, and render Ground Truth material images for each view. The details of the scene can be found in the supplementary material.

\noindent \textbf{Real dataset.} 
Widely used real datasets of large-scale scenes, \eg, Scannet\cite{scannet}, Matterport3D\cite{Matterport3D} and Replica\cite{replica19arxiv}, lack full-HDR images. Therefore, we collect 10 full-HDR real scenes. For each scene, 10 to 20 full-HDR panoramic images are captured by merging 7 bracketed exposures (from $\frac{1}{25000}$s to $\frac{1}{8}$s). 

\subsection{Baselines and Metrics}
\label{sec:exp_baseline}
To our best knowledge, there are only a few methods that recover the SVBRDFs from multi-view images for large-scale scenes. We compare with the following inverse rendering approaches: (1) The state-of-the-art single image learning-based method: PhyIR\cite{li2022phyir}; (2) The state-of-the-art multi-view object-centric neural rendering methods: InvRender\cite{zhang2022invrender}, NVDIFFREC\cite{munkberg2021nvdiffrec} and NeILF\cite{yao2022neilf}. Please note that these object-centric approaches are unsuitable to evaluate on large-scale scenes due to simple illumination representation except for NeILF. For fair comparisons, we integrate their material optimization strategies with our hybrid lighting representation, which is designed specifically for large-scale real-world scenes.~\footnote{we tried to compare with optimization-based methods~\cite{nimierdavid2021material, Haefner213dv}, but failed to get available results due to the lack of available source code.}

We use Peak Signal-to-Noise Ratio (PSNR), Structural Similarity Index Measure (SSIM)~\cite{
ssim} and Mean Squared Error (MSE) to evaluate the material predictions and the re-rendered images for quantitative comparisons. Moreover, we use Mean Absolute Error (MAE) and SSIM to evaluate the relighting images rendered by different lighting representations.

\begin{figure*}[t]
    \centering
    \begin{overpic}[scale=0.115]{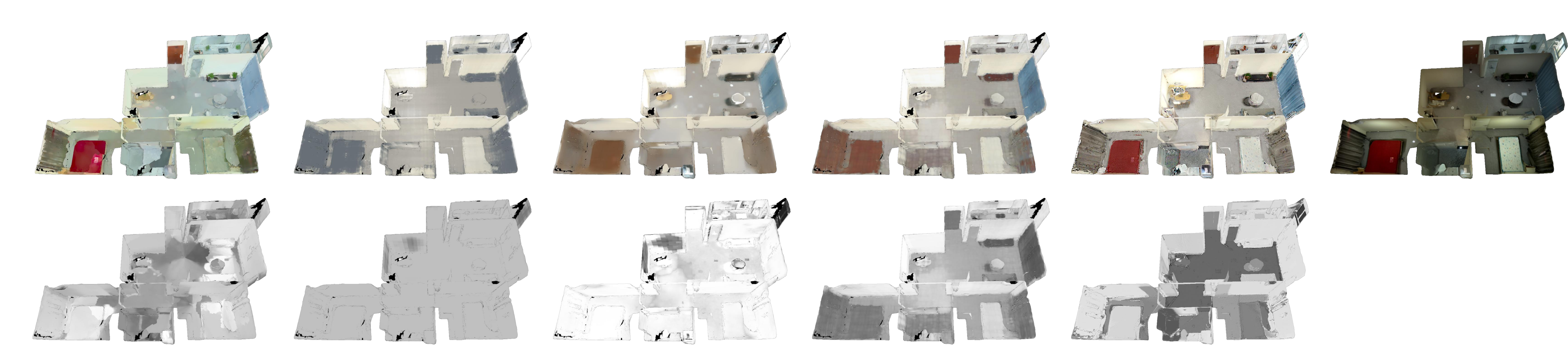}
    \put(6,21){\footnotesize PhyIR\cite{li2022phyir}}
    \put(21,21){\footnotesize InvRender\cite{zhang2022invrender}}
    \put(36.5,21){\footnotesize NVDIFFREC\cite{munkberg2021nvdiffrec}}
    \put(56,21){\footnotesize NeILF\cite{yao2022neilf}}
    \put(74,21){\footnotesize Ours}
    \put(90,21){\footnotesize RGB}
    
    \put(0,0){\rotatebox{90} {\footnotesize Roughness}}
    \put(0,12){\rotatebox{90} {\footnotesize Albedo}}
    
    \end{overpic}
    \caption{\textbf{Qualitative comparison in the 3D view on challenging real dataset.} This sample is Scene 1. Our method reconstructs globally-consistent and physically-reasonable SVBRDFs while other approaches struggle to produce consistent results and disentangle ambiguity of materials. Note that the low roughness (around 0.15 in ours) leads to the strong highlights, which are similar to GT.}
    \label{fig:comparison_real_main}
\end{figure*}

\begin{figure*}[t]
    \centering
    \begin{overpic}[scale=0.25]{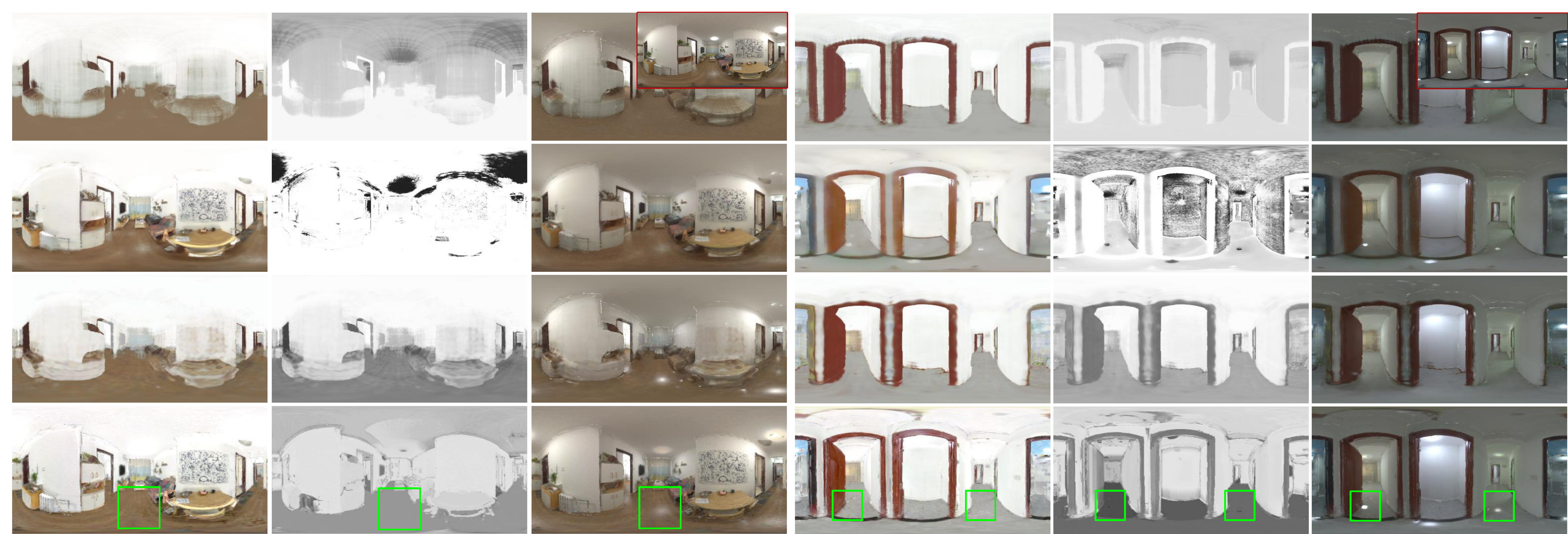}
    \put(6,33.5){\normalsize Albedo}
    \put(21,33.5){\normalsize Roughness}
    \put(37,33.5){\normalsize Rendering}
    \put(55,33.5){\normalsize Albedo}
    \put(70,33.5){\normalsize Roughness}
    \put(88,33.5){\normalsize Rendering}
    
    \put(0,3){\rotatebox{90} {\tiny Ours}}
    \put(0,10){\rotatebox{90} {\tiny NeILF\cite{yao2022neilf}}}
    \put(0,17){\rotatebox{90} {\tiny NVDIFFREC\cite{munkberg2021nvdiffrec}}}
    \put(0,26){\rotatebox{90} {\tiny InvRender\cite{zhang2022invrender}}}
    
    \end{overpic}
    \caption{\textbf{Qualitative comparison in the image view on challenging real dataset.} From left to right: Scene 8 and Scene 9. \textcolor[RGB]{200,0,0}{Red} denotes the Ground Truth image. Our physically-reasonable materials are able to render similar appearance to GT. Note that Invrender\cite{zhang2022invrender} and NeILF\cite{yao2022neilf} do not produce correct highlights, and NVDIFFREC\cite{munkberg2021nvdiffrec} fails to distinguish the ambiguity between albedo and roughness.}
    \label{fig:comparison_real2_main}
\end{figure*}
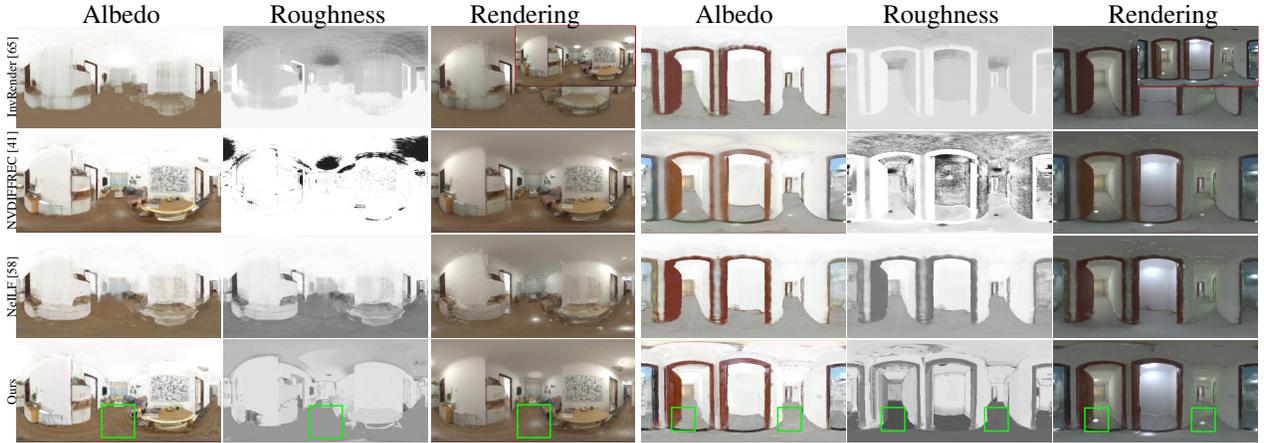

\subsection{Comparisons}
\label{sec:exp_compare}
\noindent \textbf{Evaluation on synthetic dataset.} As shown in Tab.~\ref{tb:syn} and Fig.~\ref{fig:syn_roughness}. 
Our method significantly outperforms the state-of-the-arts in roughness estimation and our roughness is able to produce physically-reasonable specular reflectance. Moveover, NeILF\cite{yao2022neilf} with our hybrid lighting representation successfully disentangles the ambiguity between material and lighting, compared to their implicit representation.

\noindent \textbf{Evaluation on real dataset.} More importantly, we conduct the experiment on our challenging real dataset containing complex materials and illumination. The quantitative comparison in Tab.~\ref{tb:real} shows our approach outperforms previous methods. Although these methods have approximate re-rendering error, only our proposed approach disentangles globally-consistent and physically-reasonable materials. We show the qualitative comparison in the 3D view in Fig.~\ref{fig:comparison_real_main} and the 2D image views in Fig.~\ref{fig:comparison_real2_main}. PhyIR\cite{li2022phyir} suffers from poor generalization performance due to a large domain gap and fails in globally-consistent predictions. InvRender\cite{zhang2022invrender}, NVDIFFREC\cite{munkberg2021nvdiffrec} and NeILF\cite{yao2022neilf} produce blur predictions with artifacts and struggle to disentangle correct materials. Although NVDIFFREC\cite{munkberg2021nvdiffrec} reaches similar performance to our method in Tab.~\ref{tb:real}, it fails to disentangle the ambiguity between albedo and roughness, \eg, highlights in the specular component are recovered incorrectly as diffuse albedo.

\begin{table}[t]
	\centering \caption{\textbf{Quantitative comparison of re-rendered images on our real dataset.} }\label{tb:real}
    \begin{tabular}{cccc}
       \toprule
       Method & PSNR$\uparrow$ & SSIM$\uparrow$ & MSE$\downarrow$ \\
       \midrule
       InvRender~\cite{zhang2022invrender} & 21.9993 & 0.7668  & 0.0065 \\
       NVDIFFREC~\cite{munkberg2021nvdiffrec} & 23.7464 & 0.8389 & 0.0044 \\
       NeILF~\cite{yao2022neilf} & 21.9260 & 0.7687 &  0.0066 \\
       Ours & \textbf{24.6093} & \textbf{0.8623}  & \textbf{0.0035} \\
       
       \bottomrule
    \end{tabular}
\end{table}

\begin{table}[t]
	\centering \caption{\textbf{Ablation study of roughness estimation on synthetic dataset.} }\label{tb:ablation_syn}
    \begin{tabular}{cccc}
       \toprule
       Method & PSNR$\uparrow$ & SSIM$\uparrow$ & MSE$\downarrow$ \\
       \midrule
       Baseline & 7.8012 & 0.52680  & 0.1659 \\
       w/o Stage \uppercase\expandafter{\romannumeral1} & 10.3561 & 0.7044 & 0.0921 \\
       w/o Stage \uppercase\expandafter{\romannumeral2} & 7.9627 & 0.5570 &  0.1599 \\
       w/o Stage \uppercase\expandafter{\romannumeral3} & 17.4177 & 0.8347 &  0.0181 \\
       Ours & \textbf{20.2132} & \textbf{0.9161}  & \textbf{0.0095} \\
       
       \bottomrule
    \end{tabular}
\end{table}

\subsection{Ablation studies}
\label{sec:exp_ablation}
To showcase the effectiveness of proposed lighting representation and material optimization strategy, we ablate the TBL, hybrid lighting representation, albedo initialization in Stage \uppercase\expandafter{\romannumeral1}, VHL-based sampling and semantics-based propagation for roughness estimation in Stage \uppercase\expandafter{\romannumeral2}, and segmentation-based fine-tuning in Stage \uppercase\expandafter{\romannumeral3}.

\begin{figure}[t]
    \centering
    \begin{overpic}[scale=0.09]{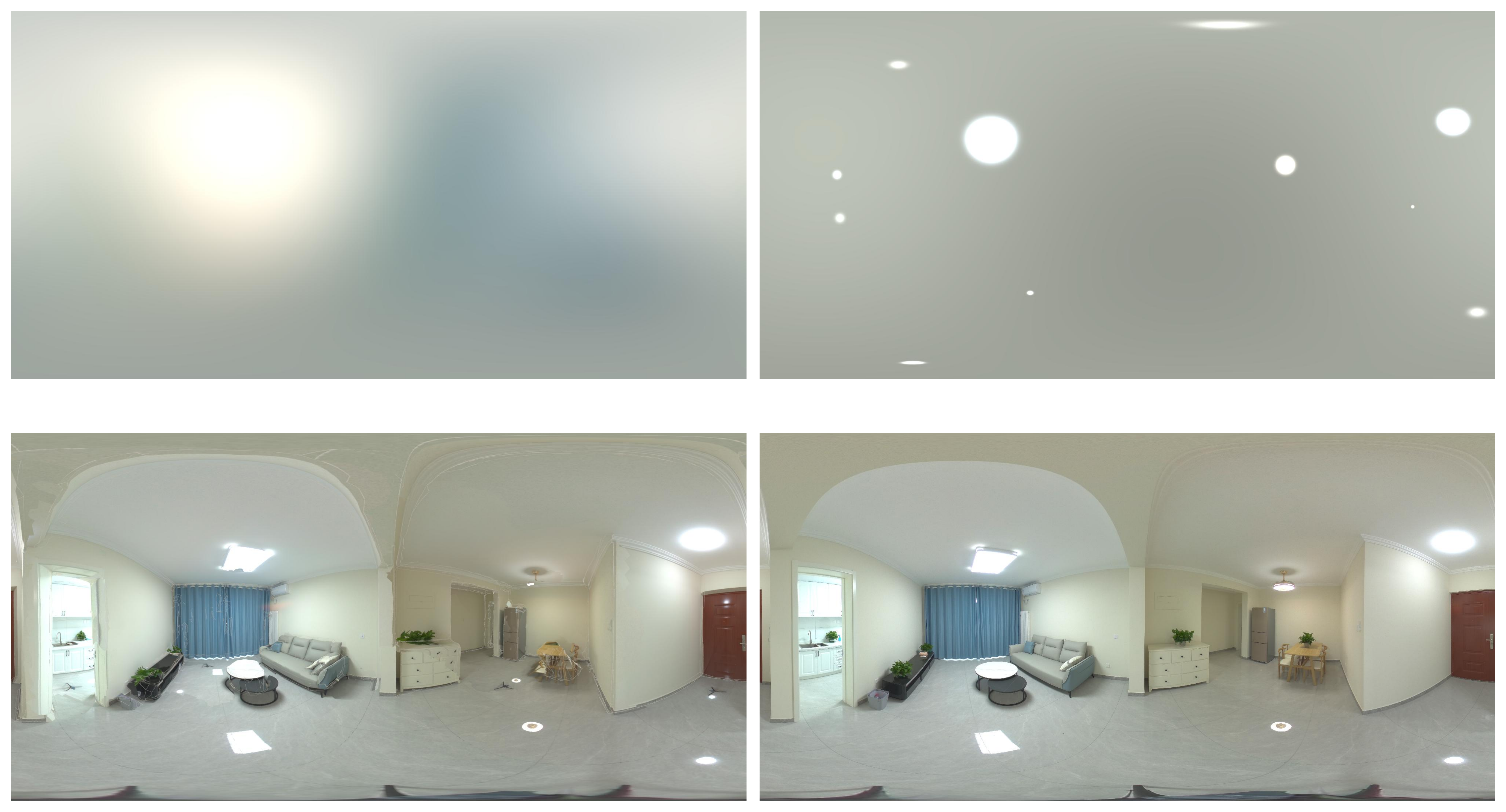}
    \put(20,26.3){\footnotesize (a) SH}
    \put(70,26.3){\footnotesize (b) SG}
    \put(19,-2.4){\footnotesize (c) TBL}
    \put(70,-2.4){\footnotesize (d) GT}
    
    \end{overpic}
    \caption{\textbf{Comparison of different lighting representations.} The result of TBL is reprojected from 3D mesh. The proposed TBL exhibits high fidelity both in low-frequency and high-frequency.}
    \label{fig:tbl_comparison}
\end{figure}

\noindent \textbf{Effectiveness of TBL.} We compare the proposed TBL to SH lighting and SG lighting widely used in previous methods\cite{Garon_2019_CVPR,Zhou_2019_ICCV,li2020inverse,wang2021learning}. As shown in Fig.~\ref{fig:tbl_comparison}, our TBL exhibits high fidelity in high-frequency features. Moreover, we evaluate the relighting error of three re-rendering virtual spheres rendered by different lighting representations in Tab.~\ref{tb:re_rendering}.
Except for accuracy, the TBL only costs around 20 MB storage to represent illumination while the dense grid-based VSG lighting \cite{wang2021learning} costs around 1 GB storage and the sparse grid-based SH lighting, Plenoxels~\cite{yu_and_fridovichkeil2021plenoxels}, costs around 750 MB storage. Therefore, our TBL achieves improved accuracy while being compact in storage.

\begin{figure}[t]
    \centering
    \begin{overpic}[scale=0.15]{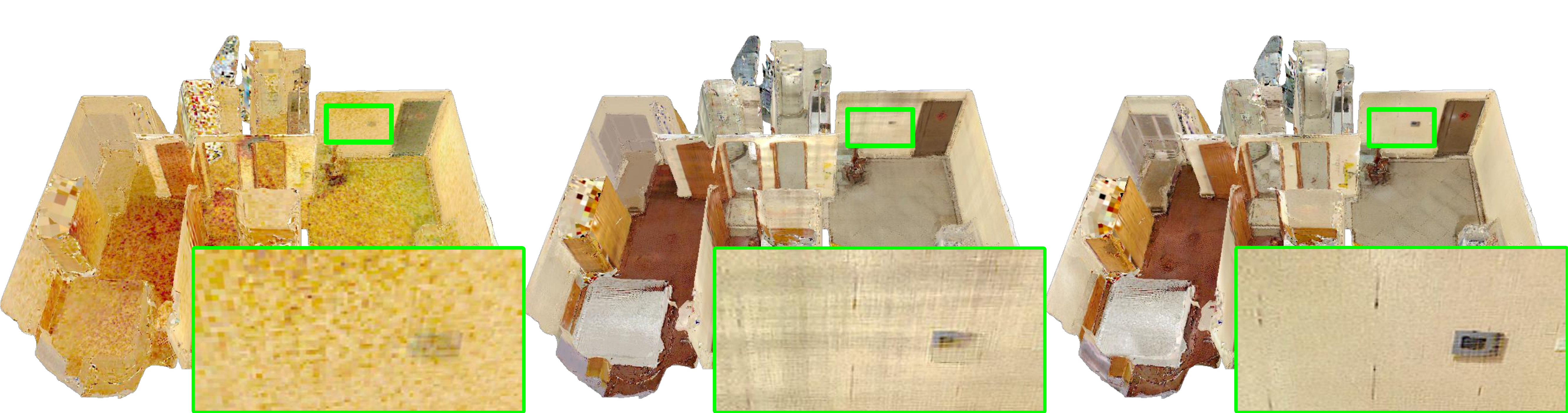}
    \put(8,26.3){\footnotesize Source TBL}
    \put(45,26.3){\footnotesize NIrF}
    \put(79.5,26.3){\footnotesize IrT}
    
    \end{overpic}
    \caption{\textbf{Ablation study of hybrid lighting representation.} This sample is Scene 11. Note that we decrease the resolution of input images and samples in Source TBL for keeping almost same time cost.}
    \label{fig:ablation_irrt}
\end{figure}

\begin{table}[t]
	\centering \caption{\textbf{Quantitative comparison of relighting spheres.} We use 5th order SH coefficients and 12 SG lobes for the comparison.}\label{tb:re_rendering}
    \scalebox{0.81}{
    \begin{tabular}{*{7}{c}}
      \toprule
      \multirow{2}*{Lighting} & \multicolumn{2}{c}{Diffuse} & \multicolumn{2}{c}{Matte Sliver} & \multicolumn{2}{c}{Mirror Sliver} \\
      \cmidrule(lr){2-3}\cmidrule(lr){4-5}\cmidrule(lr){6-7}
      & MAE$\downarrow$  & SSIM$\uparrow$  & MAE$\downarrow$ & SSIM$\uparrow$  & MAE$\downarrow$ & SSIM$\uparrow$  \\
      \midrule
      SH &0.0602&0.9982 &0.0811&0.9977 &0.1083&0.9801\\
      SG &0.0027&\textbf{0.9995} &0.0054&0.9992 &0.0348&0.9815 \\
      TBL &\textbf{0.0021}&0.9994 &\textbf{0.0028}&\textbf{0.9992} &\textbf{0.0055}&\textbf{0.9984} \\
      \bottomrule
    \end{tabular}
    }
\end{table}

\noindent \textbf{Effectiveness of Hybrid Lighting Representation.} We compare the hybrid lighting representation described in Sec.~\ref{sec:hybrid_lighting} to source TBL. As shown in Fig.~\ref{fig:ablation_irrt}, Without hybrid lighting representation, the albedo leads to noise and converges slowly. With precomputed irradiance, we can use high resolution inputs to recover detailed materials, and significantly accelerate the optimization process. The IrT produces more detailed and artifacts-free albedo, compared to the NIrF. Furthermore, we also compare to implicit lighting in Tab.~\ref{tb:syn} and Fig.~\ref{fig:syn_roughness}. NeILF~\cite{yao2022neilf} with their implicit lighting fails in disentangling the ambiguity between materials and lighting, \eg, The lighting effects are incorrectly recovered as materials in Fig.~\ref{fig:syn_roughness}.

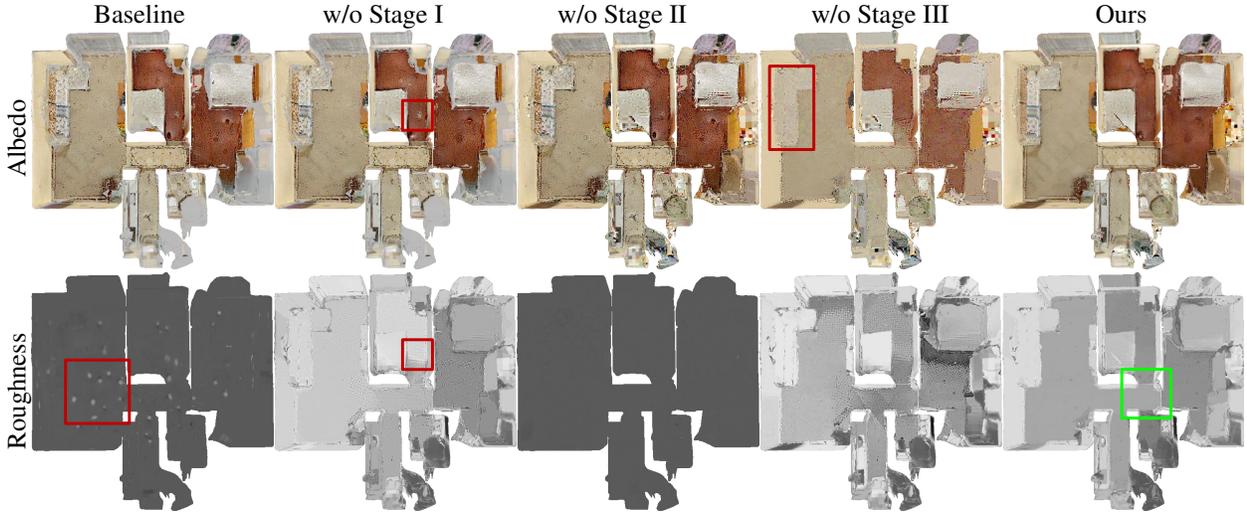
\begin{figure*}[t]
    \centering
    \begin{overpic}[scale=0.26]{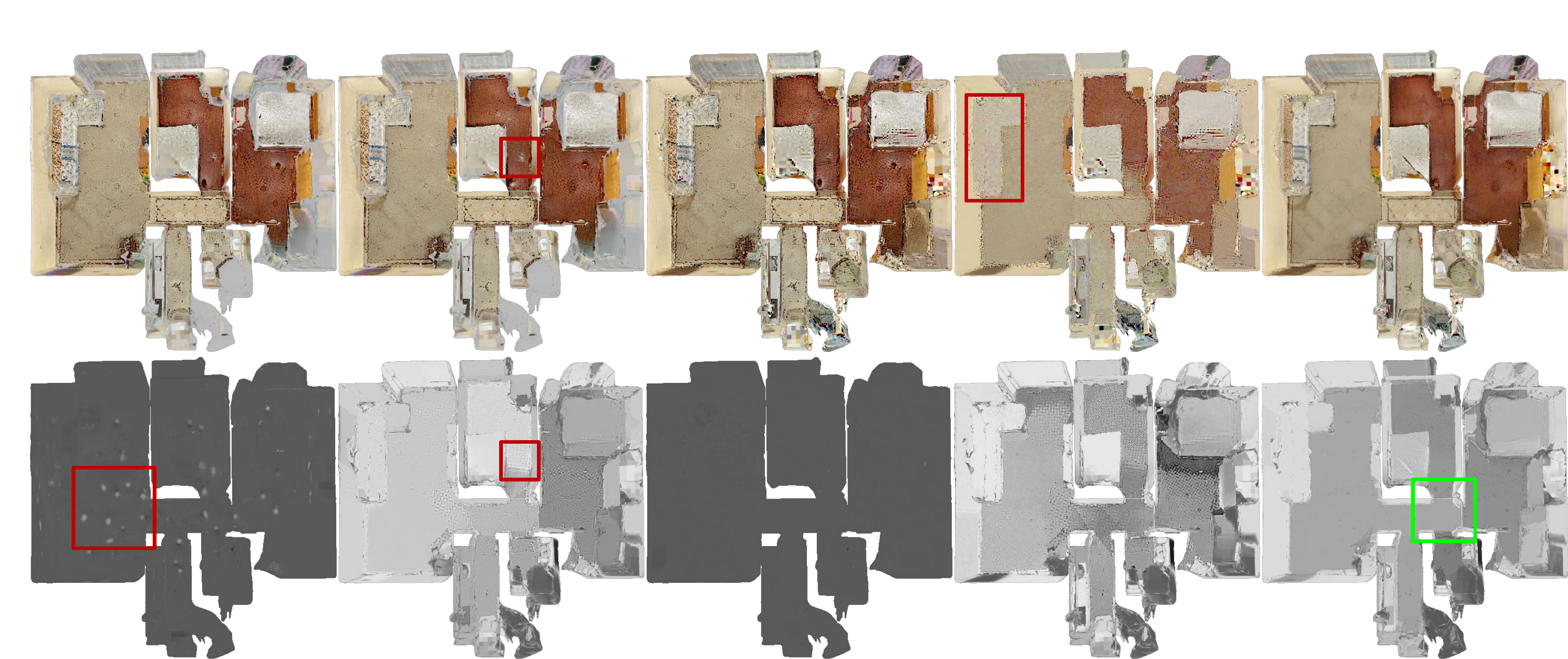}
    \put(7,39.5){\normalsize Baseline}
    \put(25.5,39.5){\normalsize w/o Stage \uppercase\expandafter{\romannumeral1}}
    \put(44.5,39.5){\normalsize w/o Stage \uppercase\expandafter{\romannumeral2}}
    \put(65,39.5){\normalsize w/o Stage \uppercase\expandafter{\romannumeral3}}
    \put(88,39.5){\normalsize Ours}
    
    \put(0,5){\rotatebox{90} {\normalsize Roughness}}
    \put(0,27){\rotatebox{90} {\normalsize Albedo}}
    
    \end{overpic}
    \caption{\textbf{Ablation study of our material optimization strategy in the 3D mesh view on challenging real dataset.} This sample is Scene 11. In baseline, we jointly optimize albedo and roughness.}
    \label{fig:ablation_material_main}
\end{figure*}

\noindent \textbf{Effectiveness of the Three Stage Strategy.} The results are shown in Tab.~\ref{tb:ablation_syn} and Fig.~\ref{fig:ablation_material_main}. The roughness of baseline fails to converge and only the highlight regions are updated. Without albedo initialization in Stage \uppercase\expandafter{\romannumeral1}, albedo in highlight regions is over-bright and leads to incorrect roughness. VHL-based sampling and semantics-based propagation in Stage \uppercase\expandafter{\romannumeral2} is crucial to recover the reasonable roughness of areas where highlights are not observed. 
Segmentation-based fine-tuning in Stage \uppercase\expandafter{\romannumeral3} produces detailed albedo, makes final roughness smoother and prevent the wrong propagation of roughness between different materials.

\subsection{Applications}
\label{sec:exp_app}
\begin{figure}[t]
    \centering
    \begin{overpic}[scale=0.36]{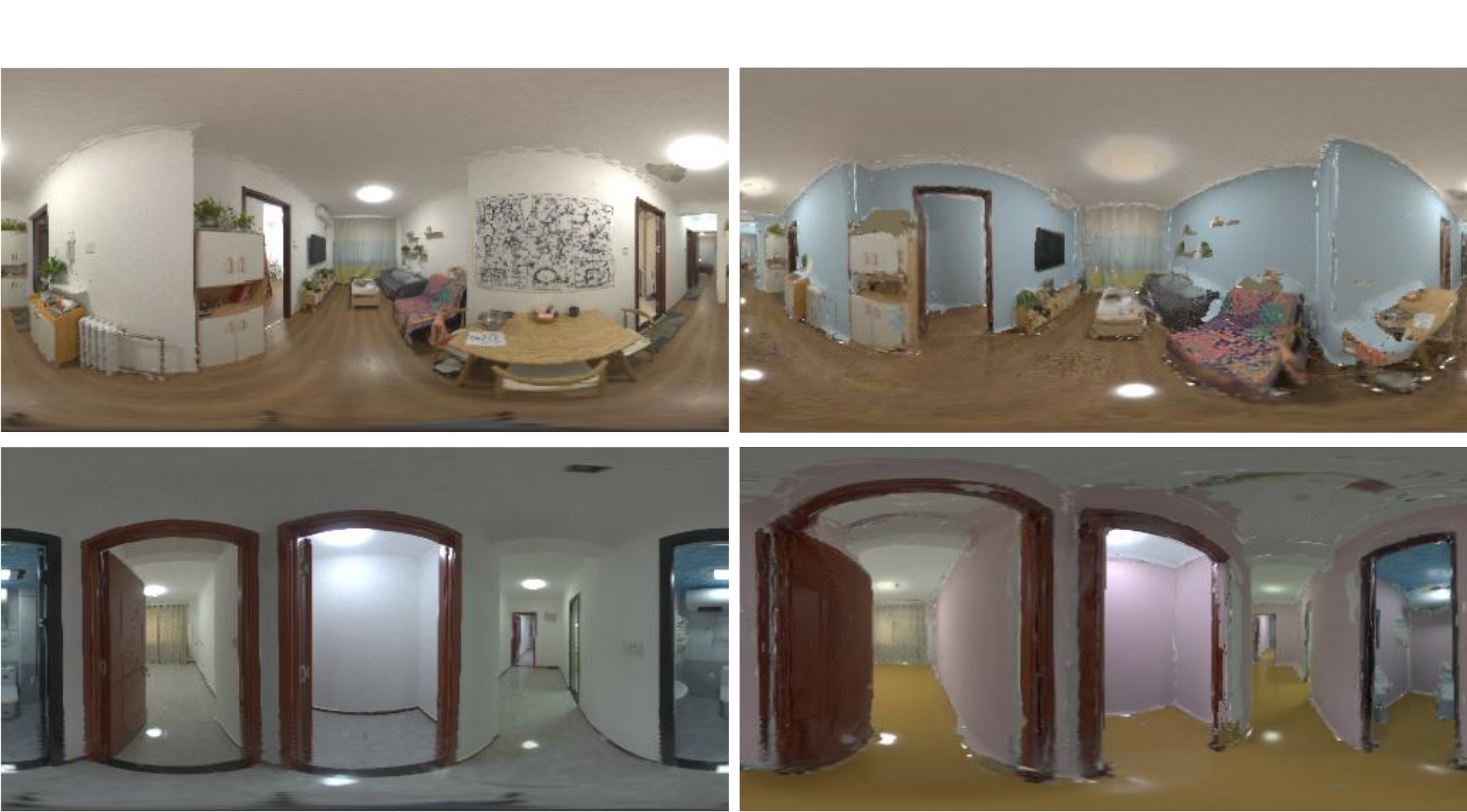}
    \put(15,52){\small Nearest View}
    \put(60,52){\small Edited Novel View}
    \put(-3,5){\rotatebox{90} {\small Scene 9}}
    \put(-3,32){\rotatebox{90} {\small Scene 8}}
    
    \end{overpic}
    \caption{\textbf{Editable novel view synthesis.} In Scene 8, we edit the albedo of the wall, and edit the roughness of the floor. In Scene 9, we edit the albedo of the floor and the wall. Our method produces convincing results (see the lighting effects in the floor and wall).}
    \label{fig:app_nv_main}
\end{figure}
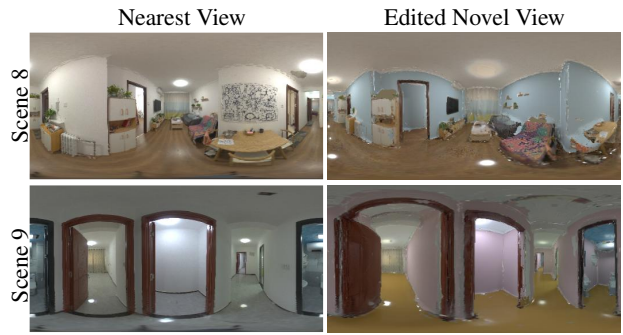

Our final output is a triangle mesh with PBR material textures, which is compatible with standard graphic engines and 3D modeling tools. We demonstrate in Fig.~\ref{fig:fig1} that the proposed approach is able to produce convincing results on material editing, editable novel view synthesis and relighting. Moreover, we show several results of editable novel view synthesis in Fig.~\ref{fig:app_nv_main}. Note that the view-dependent specular highlights reasonably change as view changes.
See the supplementary material for more results.

\section{Conclusion}
\label{sec:conclusion}

In this paper, we propose a novel inverse rendering framework that recovers globally-consistent lighting and materials from posed sparse-view images and geometry for large-scale scenes. Our texture-based lighting, which not only represents infinite-bounce global illumination but also is compact and globally-consistent, is suitable for modelling illumination of large-scale scenes. Our material optimization strategy leveraging semantics and room segmentation priors is able to reconstruct physically-reasonable and globally-consistent PBR materials. Such a triangle mesh with material textures is compatible with common graphic engines, which benefits several downstream applications such as material editing, editable novel view synthesis and relighting.

\section*{Acknowledgements}
We thank reviewers for their constructive comments, and thank Kunlong Li and Wei Li for generating annotations. This work is supported in part by the National Natural Science Foundation of China (NFSC) (No. 62002295) and Shanxi Provincial Key R\&D Program (No. 2021KWZ-03).

{\small
\bibliographystyle{ieee_fullname}
\bibliography{11_references}

\begin{thebibliography}{10}\itemsep=-1pt

\bibitem{azinovic2019inverse}
Dejan Azinovic, Tzu{-}Mao Li, Anton Kaplanyan, and Matthias Nie{\ss}ner.
\newblock Inverse path tracing for joint material and lighting estimation.
\newblock In {\em Proceedings of the IEEE/CVF Conference on Computer Vision and
  Pattern Recognition.}, 2019.

\bibitem{barron2013intrinsic}
Jonathan~T Barron and Jitendra Malik.
\newblock Intrinsic scene properties from a single rgb-d image.
\newblock In {\em Proceedings of the IEEE/CVF Conference on Computer Vision and
  Pattern Recognition.}, pages 17--24, 2013.

\bibitem{SIRFS}
Jonathan~T. Barron and Jitendra Malik.
\newblock Shape, illumination, and reflectance from shading.
\newblock {\em IEEE Transactions on Pattern Analysis and Machine
  Intelligence.}, 37(8):1670--1687, 2015.

\bibitem{iiwdataset}
Sean Bell, Kavita Bala, and Noah Snavely.
\newblock Intrinsic images in the wild.
\newblock {\em ACM Transactions on Graphics.}, 33(4), 2014.

\bibitem{bisai18intrinsic}
Sai Bi, Nima~Khademi Kalantari, and Ravi Ramamoorthi.
\newblock {Deep Hybrid Real and Synthetic Training for Intrinsic
  Decomposition}.
\newblock In Wenzel Jakob and Toshiya Hachisuka, editors, {\em Eurographics
  Symposium on Rendering.} The Eurographics Association, 2018.

\bibitem{bi2020nrf}
Sai Bi, Zexiang Xu, Pratul~P. Srinivasan, Ben Mildenhall, Kalyan Sunkavalli,
  Milos Hasan, Yannick Hold{-}Geoffroy, David~J. Kriegman, and Ravi
  Ramamoorthi.
\newblock Neural reflectance fields for appearance acquisition.
\newblock 2020.

\bibitem{boss2021nerd}
Mark Boss, Raphael Braun, Varun Jampani, Jonathan~T. Barron, Ce Liu, and
  Hendrik~P.A. Lensch.
\newblock Nerd: Neural reflectance decomposition from image collections.
\newblock In {\em Proceedings of the IEEE International Conference on Computer
  Vision.}, 2021.

\bibitem{TwoShotShapeAndBrdf}
Mark Boss, Varun Jampani, Kihwan Kim, Hendrik~P.A. Lensch, and Jan Kautz.
\newblock Two-shot spatially-varying brdf and shape estimation.
\newblock In {\em Proceedings of the IEEE/CVF Conference on Computer Vision and
  Pattern Recognition.}, 2020.

\bibitem{Burley2012pbrdisney}
Brent Burley.
\newblock Physically-based shading at disney.
\newblock 2012.

\bibitem{Matterport3D}
Angel Chang, Angela Dai, Thomas Funkhouser, Maciej Halber, Matthias Niessner,
  Manolis Savva, Shuran Song, Andy Zeng, and Yinda Zhang.
\newblock Matterport3d: Learning from rgb-d data in indoor environments.
\newblock {\em International Conference on 3D Vision.}, 2017.

\bibitem{mmseg2020}
MMSegmentation Contributors.
\newblock {MMSegmentation}: Openmmlab semantic segmentation toolbox and
  benchmark.
\newblock \url{https://github.com/open-mmlab/mmsegmentation}, 2020.

\bibitem{scannet}
Angela Dai, Angel~X. Chang, Manolis Savva, Maciej Halber, Thomas Funkhouser,
  and Matthias Nie{\ss}ner.
\newblock Scannet: Richly-annotated 3d reconstructions of indoor scenes.
\newblock In {\em Proceedings of the IEEE/CVF Conference on Computer Vision and
  Pattern Recognition.}, 2017.

\bibitem{debevec2008rendering}
Paul Debevec.
\newblock Rendering synthetic objects into real scenes: Bridging traditional
  and image-based graphics with global illumination and high dynamic range
  photography.
\newblock In {\em ACM SIGGRAPH 2008 classes}, page~32, 2008.

\bibitem{multiSVBRDF2019}
Valentin Deschaintre, Miika Aittala, Fr\'edo Durand, George Drettakis, and
  Adrien Bousseau.
\newblock Flexible svbrdf capture with a multi-image deep network.
\newblock {\em Computer Graphics Forum.}, 38(4), July 2019.

\bibitem{Elfes1989UsingOG}
Alberto Elfes.
\newblock Using occupancy grids for mobile robot perception and navigation.
\newblock {\em Computer}, 22:46--57, 1989.

\bibitem{gardner2019deep}
Marc-Andr{\'e} Gardner, Yannick Hold-Geoffroy, Kalyan Sunkavalli, Christian
  Gagn{\'e}, and Jean-Fran{\c{c}}ois Lalonde.
\newblock Deep parametric indoor lighting estimation.
\newblock In {\em Proceedings of the IEEE International Conference on Computer
  Vision.}, pages 7175--7183, 2019.

\bibitem{gardner-sigasia-17}
Marc-Andr\'{e} Gardner, Kalyan Sunkavalli, Ersin Yumer, Xiaohui Shen, Emiliano
  Gambaretto, Christian Gagn\'{e}, and Jean-Fran\c{c}ois Lalonde.
\newblock Learning to predict indoor illumination from a single image.
\newblock {\em ACM Transactions on Graphics. (Proceedings of SIGGRAPH Asia.)},
  9(4), 2017.

\bibitem{Garon_2019_CVPR}
Mathieu Garon, Kalyan Sunkavalli, Sunil Hadap, Nathan Carr, and Jean-Francois
  Lalonde.
\newblock Fast spatially-varying indoor lighting estimation.
\newblock In {\em Proceedings of the IEEE/CVF Conference on Computer Vision and
  Pattern Recognition.}, June 2019.

\bibitem{Greger1998TheIV}
Gene Greger, Peter Shirley, Philip~M. Hubbard, and Donald~P. Greenberg.
\newblock The irradiance volume.
\newblock {\em IEEE Computer Graphics and Applications}, 18:32--43, 1998.

\bibitem{Haefner213dv}
B. Haefner, S. Green, A. Oursland, D. Andersen, M. Goesele, D. Cremers, R.
  Newcombe, and T. Whelan.
\newblock Recovering real-world reflectance properties and shading from {HDR}
  imagery.
\newblock In {\em International Conference on 3D Vision.}, London, UK, December
  2021.

\bibitem{hu2020jittor}
Shi-Min Hu, Dun Liang, Guo-Ye Yang, Guo-Wei Yang, and Wen-Yang Zhou.
\newblock Jittor: a novel deep learning framework with meta-operators and
  unified graph execution.
\newblock {\em Science China Information Sciences}, 63(222103):1--222103, 2020.

\bibitem{jiang2021unifuse}
Hualie Jiang, Zhe Sheng, Siyu Zhu, Zilong Dong, and Rui Huang.
\newblock Unifuse: Unidirectional fusion for 360$^{\circ}$ panorama depth
  estimation.
\newblock {\em IEEE Robotics and Automation Letters.}, 2021.

\bibitem{kajiya1986rendering}
James~T Kajiya.
\newblock The rendering equation.
\newblock In {\em ACM Transactions on Graphics.}, volume~20, pages 143--150,
  1986.

\bibitem{jointSVBRDF2019}
Kaizhang Kang, Cihui Xie, Chengan He, Mingqi Yi, Minyi Gu, Zimin Chen, Kun
  Zhou, and Hongzhi Wu.
\newblock Learning efficient illumination multiplexing for joint capture of
  reflectance and shape.
\newblock {\em ACM Transactions on Graphics.}, 38(6), Nov. 2019.

\bibitem{realshading}
Brian Karis and Epic Games.
\newblock Real shading in unreal engine 4, 2013.

\bibitem{michael2006poissonrecons}
Michael Kazhdan, Matthew Bolitho, and Hugues Hoppe.
\newblock Poisson surface reconstruction.
\newblock In {\em Proceedings of the Fourth Eurographics Symposium on Geometry
  Processing}, page 61–70, 2006.

\bibitem{Adam}
Diederik~P. Kingma and Jimmy Ba.
\newblock Adam: A method for stochastic optimization, 2015.

\bibitem{Laine2020diffrast}
Samuli Laine, Janne Hellsten, Tero Karras, Yeongho Seol, Jaakko Lehtinen, and
  Timo Aila.
\newblock Modular primitives for high-performance differentiable rendering.
\newblock {\em ACM Transactions on Graphics.}, 39(6), 2020.

\bibitem{li2022shape}
Junxuan Li and Hongdong Li.
\newblock Neural reflectance for shape recovery with shadow handling.
\newblock In {\em Proceedings of the IEEE/CVF Conference on Computer Vision and
  Pattern Recognition.}, pages 16221--16230, 2022.

\bibitem{li2021lighting}
Junxuan Li, Hongdong Li, and Yasuyuki Matsushita.
\newblock Lighting, reflectance and geometry estimation from 360° panoramic
  stereo.
\newblock In {\em Proceedings of the IEEE/CVF Conference on Computer Vision and
  Pattern Recognition.}, 2021.

\bibitem{li2018redner}
Tzu-Mao Li, Miika Aittala, Fr{\'e}do Durand, and Jaakko Lehtinen.
\newblock Differentiable monte carlo ray tracing through edge sampling.
\newblock {\em ACM Transactions on Graphics. (Proceedings of SIGGRAPH Asia.)},
  37(6):222:1--222:11, 2018.

\bibitem{li2020inverse}
Zhengqin Li, Mohammad Shafiei, Ravi Ramamoorthi, Kalyan Sunkavalli, and
  Manmohan Chandraker.
\newblock Inverse rendering for complex indoor scenes: Shape, spatially-varying
  lighting and svbrdf from a single image.
\newblock In {\em Proceedings of the IEEE/CVF Conference on Computer Vision and
  Pattern Recognition.}, pages 2475--2484, 2020.

\bibitem{li2022editing}
Zhengqin Li, Jia Shi, Sai Bi, Rui Zhu, Kalyan Sunkavalli, Miloš Hašan,
  Zexiang Xu, Ravi Ramamoorthi, and Manmohan Chandraker.
\newblock Physically-based editing of indoor scene lighting from a single
  image.
\newblock In {\em Proceedings of the European Conference on Computer Vision.},
  2022.

\bibitem{li2018cgintrinsics}
Zhengqi Li and Noah Snavely.
\newblock Cgintrinsics: Better intrinsic image decomposition through
  physically-based rendering.
\newblock In {\em Proceedings of the European Conference on Computer Vision.},
  2018.

\bibitem{li2022phyir}
Zhen Li, Lingli Wang, Xiang Huang, Cihui Pan, and Jiaqi. Yang.
\newblock Phyir: Physics-based inverse rendering for panoramic indoor images.
\newblock In {\em Proceedings of the IEEE/CVF Conference on Computer Vision and
  Pattern Recognition.}, 2022.

\bibitem{li2018learning}
Zhengqin Li, Zexiang Xu, Ravi Ramamoorthi, Kalyan Sunkavalli, and Manmohan
  Chandraker.
\newblock Learning to reconstruct shape and spatially-varying reflectance from
  a single image.
\newblock In {\em ACM Transactions on Graphics. (Proceedings of SIGGRAPH
  Asia.)}, page 269, 2018.

\bibitem{openrooms}
Zhengqin Li, Ting-Wei Yu, Shen Sang, Sarah Wang, Meng Song, Yuhan Liu, Yu-Ying
  Yeh, Rui Zhu, Nitesh Gundavarapu, Jia Shi, Sai Bi, Hong-Xing Yu, Zexiang Xu,
  Kalyan Sunkavalli, Milos Hasan, Ravi Ramamoorthi, and Manmohan Chandraker.
\newblock Openrooms: An open framework for photorealistic indoor scene
  datasets.
\newblock In {\em Proceedings of the IEEE/CVF Conference on Computer Vision and
  Pattern Recognition.}, pages 7190--7199, June 2021.

\bibitem{liu2019softras}
Shichen Liu, Tianye Li, Weikai Chen, and Hao Li.
\newblock Soft rasterizer: A differentiable renderer for image-based 3d
  reasoning.
\newblock {\em Proceedings of the IEEE International Conference on Computer
  Vision.}, Oct 2019.

\bibitem{Luan2021UnifiedSA}
Fujun Luan, Shuang Zhao, Kavita Bala, and Zhao Dong.
\newblock Unified shape and svbrdf recovery using differentiable monte carlo
  rendering.
\newblock {\em Computer Graphics Forum.}, 40, 2021.

\bibitem{mildenhall2020nerf}
Ben Mildenhall, Pratul~P. Srinivasan, Matthew Tancik, Jonathan~T. Barron, Ravi
  Ramamoorthi, and Ren Ng.
\newblock Nerf: Representing scenes as neural radiance fields for view
  synthesis.
\newblock In {\em Proceedings of the European Conference on Computer Vision.},
  2020.

\bibitem{munkberg2021nvdiffrec}
Jacob Munkberg, Jon Hasselgren, Tianchang Shen, Jun Gao, Wenzheng Chen, Alex
  Evans, Thomas Mueller, and Sanja Fidler.
\newblock Extracting triangular 3d models, materials, and lighting from images.
\newblock In {\em Proceedings of the IEEE/CVF Conference on Computer Vision and
  Pattern Recognition.}, 2022.

\bibitem{MobileSVBRDF2018}
Giljoo Nam, Joo~Ho Lee, Diego Gutierrez, and Min~H. Kim.
\newblock Practical svbrdf acquisition of 3d objects with unstructured flash
  photography.
\newblock {\em ACM Transactions on Graphics. (Proceedings of SIGGRAPH Asia.)},
  37(6):267:1--12, 2018.

\bibitem{nestmeyer2020faceRelighting}
Thomas Nestmeyer, Jean-FranÃ§ois Lalonde, Iain Matthews, and Andreas~M
  Lehrmann.
\newblock Learning physics-guided face relighting under directional light.
\newblock In {\em Proceedings of the IEEE/CVF Conference on Computer Vision and
  Pattern Recognition.}, 2020.

\bibitem{nimierdavid2021material}
Merlin Nimier-David, Zhao Dong, Wenzel Jakob, and Anton Kaplanyan.
\newblock {Material and Lighting Reconstruction for Complex Indoor Scenes with
  Texture-space Differentiable Rendering}.
\newblock In {\em Eurographics Symposium on Rendering.}, 2021.

\bibitem{roberts2021hypersim}
Mike Roberts, Jason Ramapuram, Anurag Ranjan, Atulit Kumar, Miguel~Angel
  Bautista, Nathan Paczan, Russ Webb, and Joshua~M. Susskind.
\newblock {Hypersim}: {A} photorealistic synthetic dataset for holistic indoor
  scene understanding.
\newblock In {\em Proceedings of the IEEE International Conference on Computer
  Vision.}, 2021.

\bibitem{yu_and_fridovichkeil2021plenoxels}
{Sara Fridovich-Keil and Alex Yu}, Matthew Tancik, Qinhong Chen, Benjamin
  Recht, and Angjoo Kanazawa.
\newblock Plenoxels: Radiance fields without neural networks.
\newblock In {\em Proceedings of the IEEE/CVF Conference on Computer Vision and
  Pattern Recognition.}, 2022.

\bibitem{Schmitt_2020_CVPR}
Carolin Schmitt, Simon Donne, Gernot Riegler, Vladlen Koltun, and Andreas
  Geiger.
\newblock On joint estimation of pose, geometry and svbrdf from a handheld
  scanner.
\newblock In {\em Proceedings of the IEEE/CVF Conference on Computer Vision and
  Pattern Recognition.}, June 2020.

\bibitem{colmap}
Johannes~Lutz Sch\"{o}nberger, Enliang Zheng, Marc Pollefeys, and Jan-Michael
  Frahm.
\newblock Pixelwise view selection for unstructured multi-view stereo.
\newblock In {\em Proceedings of the European Conference on Computer Vision.},
  2016.

\bibitem{neuralSengupta19}
Soumyadip Sengupta, Jinwei Gu, Kihwan Kim, Guilin Liu, David~W. Jacobs, and Jan
  Kautz.
\newblock Neural inverse rendering of an indoor scene from a single image.
\newblock In {\em Proceedings of the IEEE International Conference on Computer
  Vision.}, 2019.

\bibitem{nerv2021}
Pratul~P. Srinivasan, Boyang Deng, Xiuming Zhang, Matthew Tancik, Ben
  Mildenhall, and Jonathan~T. Barron.
\newblock Nerv: Neural reflectance and visibility fields for relighting and
  view synthesis.
\newblock In {\em Proceedings of the IEEE/CVF Conference on Computer Vision and
  Pattern Recognition.}, 2021.

\bibitem{replica19arxiv}
Julian Straub, Thomas Whelan, Lingni Ma, Yufan Chen, Erik Wijmans, Simon Green,
  Jakob~J. Engel, Raul Mur-Artal, Carl Ren, Shobhit Verma, Anton Clarkson,
  Mingfei Yan, Brian Budge, Yajie Yan, Xiaqing Pan, June Yon, Yuyang Zou,
  Kimberly Leon, Nigel Carter, Jesus Briales, Tyler Gillingham, Elias Mueggler,
  Luis Pesqueira, Manolis Savva, Dhruv Batra, Hauke~M. Strasdat, Renzo~De
  Nardi, Michael Goesele, Steven Lovegrove, and Richard Newcombe.
\newblock The {R}eplica dataset: A digital replica of indoor spaces.
\newblock {\em arXiv preprint arXiv:1906.05797}, 2019.

\bibitem{wang2022r2l}
Huan Wang, Jian Ren, Zeng Huang, Kyle Olszewski, Menglei Chai, Yun Fu, and
  Sergey Tulyakov.
\newblock R2l: Distilling neural radiance field to neural light field for
  efficient novel view synthesis.
\newblock In {\em Proceedings of the European Conference on Computer Vision.},
  2022.

\bibitem{ssim}
Zhou Wang, A.C. Bovik, H.R. Sheikh, and E.P. Simoncelli.
\newblock Image quality assessment: from error visibility to structural
  similarity.
\newblock {\em IEEE Transactions on Image Processing.}, 13(4):600--612, 2004.

\bibitem{wang2021learning}
Zian Wang, Jonah Philion, Sanja Fidler, and Jan Kautz.
\newblock Learning indoor inverse rendering with 3d spatially-varying lighting.
\newblock In {\em Proceedings of the IEEE International Conference on Computer
  Vision.}, 2021.

\bibitem{di88}
Gregory~J. Ward, Francis~M. Rubinstein, and Robert~D. Clear.
\newblock A ray tracing solution for diffuse interreflection.
\newblock {\em Computer Graphics}, 22(4), 1988.

\bibitem{wood00slf}
Daniel~N. Wood, Daniel~I. Azuma, Ken Aldinger, Brian Curless, Tom Duchamp,
  David~H. Salesin, and Werner Stuetzle.
\newblock Surface light fields for 3d photography.
\newblock SIGGRAPH '00, page 287–296, 2000.

\bibitem{yang2022psnerf}
Wenqi Yang, Guanying Chen, Chaofeng Chen, Zhenfang Chen, and Kwan-Yee~K. Wong.
\newblock Ps-nerf: Neural inverse rendering for multi-view photometric stereo.
\newblock In {\em Proceedings of the European Conference on Computer Vision.},
  2022.

\bibitem{yao2022neilf}
Yao Yao, Jingyang Zhang, Jingbo Liu, Yihang Qu, Tian Fang, David McKinnon,
  Yanghai Tsin, and Long Quan.
\newblock Neilf: Neural incident light field for physically-based material
  estimation.
\newblock In {\em Proceedings of the European Conference on Computer Vision.},
  2022.

\bibitem{yariv2020multiview}
Lior Yariv, Yoni Kasten, Dror Moran, Meirav Galun, Matan Atzmon, Basri Ronen,
  and Yaron Lipman.
\newblock Multiview neural surface reconstruction by disentangling geometry and
  appearance.
\newblock {\em Advances in Neural Information Processing Systems.}, 33, 2020.

\bibitem{yu99sig}
Yizhou Yu, Paul Debevec, Jitendra Malik, and Tim Hawkins.
\newblock Inverse global illumination: Recovering reflectance models of real
  scenes from photographs.
\newblock In {\em Proceedings of the 26th Annual Conference on Computer
  Graphics and Interactive Techniques}, SIGGRAPH '99, page 215–224, 1999.

\bibitem{zhang2016emptying}
Edward Zhang, Michael~F. Cohen, and Brian Curless.
\newblock Emptying, refurnishing, and relighting indoor spaces.
\newblock {\em ACM Transactions on Graphics. (Proceedings of SIGGRAPH Asia.)},
  35(6), 2016.

\bibitem{iron-2022}
Kai Zhang, Fujun Luan, Zhengqi Li, and Noah Snavely.
\newblock Iron: Inverse rendering by optimizing neural sdfs and materials from
  photometric images.
\newblock In {\em Proceedings of the IEEE/CVF Conference on Computer Vision and
  Pattern Recognition.}, 2022.

\bibitem{physg2020}
Kai Zhang, Fujun Luan, Qianqian Wang, Kavita Bala, and Noah Snavely.
\newblock Physg: Inverse rendering with spherical gaussians for physics-based
  material editing and relighting.
\newblock In {\em Proceedings of the IEEE/CVF Conference on Computer Vision and
  Pattern Recognition.}, 2021.

\bibitem{nerfactor}
Xiuming Zhang, Pratul~P Srinivasan, Boyang Deng, Paul Debevec, William~T
  Freeman, and Jonathan~T Barron.
\newblock {NeRFactor: Neural Factorization of Shape and Reflectance Under an
  Unknown Illumination}.
\newblock {\em ACM Transactions on Graphics.}, 2021.

\bibitem{zhang2022invrender}
Yuanqing Zhang, Jiaming Sun, Xingyi He, Huan Fu, Rongfei Jia, and Xiaowei Zhou.
\newblock Modeling indirect illumination for inverse rendering.
\newblock In {\em Proceedings of the IEEE/CVF Conference on Computer Vision and
  Pattern Recognition.}, 2022.

\bibitem{Zhou_2019_ICCV}
Hao Zhou, Xiang Yu, and David~W. Jacobs.
\newblock Glosh: Global-local spherical harmonics for intrinsic image
  decomposition.
\newblock In {\em Proceedings of the IEEE International Conference on Computer
  Vision.}, October 2019.

\bibitem{irisformer2022}
Rui Zhu, Zhengqin Li, Janarbek Matai, Fatih Porikli, and Manmohan Chandraker.
\newblock Irisformer: Dense vision transformers for single-image inverse
  rendering in indoor scenes.
\newblock In {\em Proceedings of the IEEE/CVF Conference on Computer Vision and
  Pattern Recognition.}, pages 2822--2831, June 2022.

\end{thebibliography}
}

\clearpage \appendix
\label{sec:appendix}

In this supplementary material, we provide more details of implementation (Sec.~\ref{sec:implementation}), proposed datasets (Sec.~\ref{sec:dataset}), additional experimental results (Sec.~\ref{sec:experiment}) and discussions (Sec.~\ref{sec:discussion}).


\section{Details of Implementation}\label{sec:implementation}
\subsection{BRDF Model}
In Sec. 3.2 in the main paper, $f_d$ and $f_s$ are defined as:
\begin{equation}
    f_d = \frac{A}{\pi} , f_s = \frac{DFG}{4(n\cdot v)(n \cdot l)}
\end{equation}
where $A$ is albedo; $l$ denotes light direction; $n$ denotes normal; $v$ denotes view direction; $D$ denotes Normal Distribution Function (NDF); $F$ denotes Fresnel function and $G$ is the Geometry Factor. We adopt a simplified $D$, $F$ and $G$~\cite{realshading,li2022phyir}.

The specular $D$:
\begin{equation}\label{eq:D}
\begin{aligned}
    &D = \frac{\alpha^2}{\pi {({(n \cdot h)}^2 (\alpha^2-1) +1)}^2 } ,\\
    &h = bisector(v,l) ,\\
    &\alpha = R^2 .
\end{aligned}
\end{equation}

The specular $F$:
\begin{equation}
\begin{aligned}
    &F = 0.04 + (1-0.04)2^{(-5.55473(v \cdot h) - 6.98316)(v \cdot h)} \\
\end{aligned}
\end{equation}

The specular $G$:
\begin{equation}
\begin{aligned}
    &G = G_1(l)G_1(v) ,\\
    &G_1(v) = \frac{n \cdot v}{(n \cdot v)(1-k)+k} ,\\
    &k = \frac{(R+1)^2}{8} .
\end{aligned}
\end{equation}

\begin{figure}[t]
    \centering
    \begin{overpic}[scale=0.33]{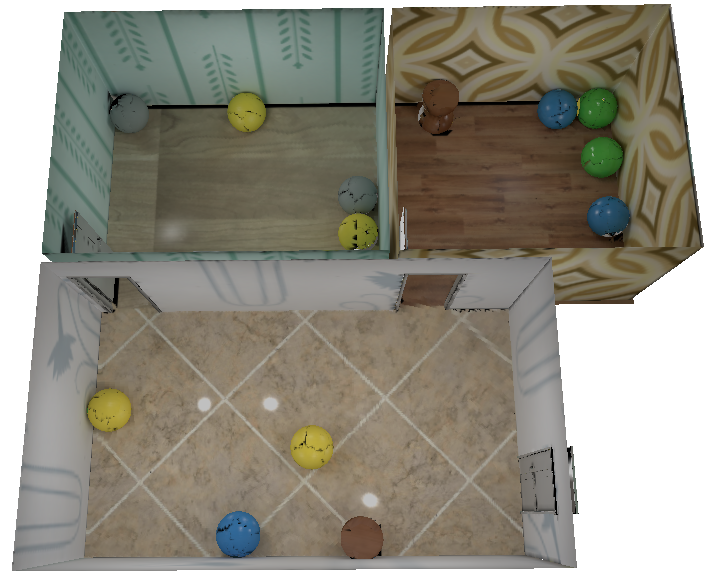}
    
    \end{overpic}
    \caption{\textbf{Overview of our synthetic dataset.} It contains diverse materials and objects.}
    \label{fig:syn_data_overview}
\end{figure}

\begin{table}[t]
	\centering \caption{\textbf{Comparison of costs.} $N$ denotes the number of images. Our method achieves competitive performance on costs compared to the highly efficient method, NVDIFFREC\cite{munkberg2021nvdiffrec}. The performance of TSDR$*$~\cite{nimierdavid2021material} is reported by their paper.}\label{tb:supp_cost}
    \begin{tabular}{ccc}
       \toprule
       Method & Time (s) & Memory (MB) \\
       \midrule
       TSDR$*$\cite{nimierdavid2021material} & 43200 & - \\
       InvRender\cite{zhang2022invrender} & 50$\times N$ & 5547\\
       NVDIFFREC\cite{munkberg2021nvdiffrec} & 42$\times N$ & \textbf{2159} \\
       NeILF$*$\cite{yao2022neilf} & 144$\times N$ & $>32510$ \\
       NeILF\cite{yao2022neilf} & 80$\times N$ & 9783\\
       Ours & \textbf{41$\times N$} & 2543  \\
       
       \bottomrule
    \end{tabular}
\end{table}

\subsection{Implementation}
We use neural networks to predict the depth image~\cite{jiang2021unifuse} and semantic segmentation~\cite{mmseg2020} for each input image. The 3D mesh of whole scene is reconstructed with depth images and poisson surface reconstruction algorithm \cite{michael2006poissonrecons}. The room segmentation is calculated by occupancy grid~\cite{Elfes1989UsingOG}.

We use 2048 samples to precompute the irradiance of sampled surface points. 
The NIrF is trained for 2000 epochs with the batch size of 16 and the total size of 1024 and we use the Adam optimizer\cite{Adam} with a learning rate of 1e-4. The resolution of IrT is 1024$\times$1024.

In material estimation, we use the Adam optimizer\cite{Adam} with a learning rate of 3e-2 for 40 epochs in all three stages. We set $\beta_{ssa}$ as 10 in stage 1, set $\beta_{sp}$ as 1 in stage 2 and set $\beta_{ssr}$ as 0.1 in stage 3. 
The resolution of albedo texture to be optimized is 2048$\times$2048 and the resolution of roughness texture to be optimized is 4096$\times$4096.
We use 16 samples to re-render the specular component in material estimation. Considering the efficiency of optimization and  the natural global illumination of proposed TBL, we apply nvdiffrast\cite{Laine2020diffrast} with deferred shading to backward the gradient of image-space materials into corresponding textures. We note that nvdiffrast is orthogonal to our pipeline, which can be replaced by other differentiable renderers\cite{li2018redner, liu2019softras, hu2020jittor}. 
The pre-computed IrT takes around 10 minutes and the optimization process of material takes around 20 minutes.

\begin{figure*}[t]
    \centering
    \begin{overpic}[scale=0.17]{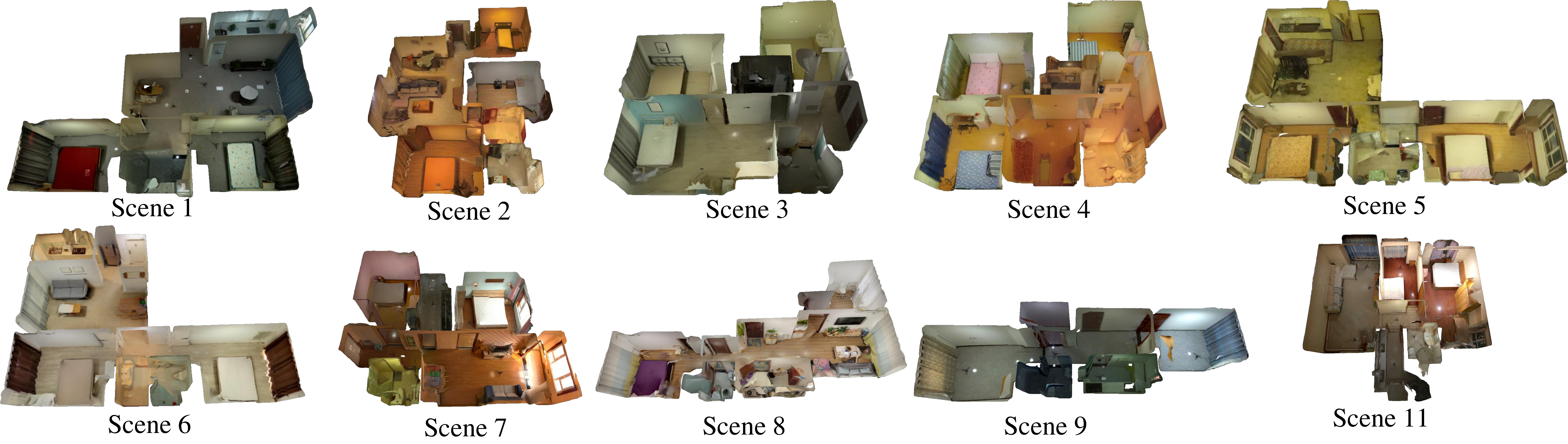}
    
    
    
    \end{overpic}
    \caption{\textbf{Overview of our challenging real dataset.} Our dataset consists of 10 Full-HDR indoor scenes with extremely complex lighting, geometry and materials.}
    \label{fig:real_data_overview}
\end{figure*}

\begin{table*}[t]
	\centering \caption{\textbf{Detailed quantitative comparison on our challenging real dataset. Although NVDIFFREC\cite{munkberg2021nvdiffrec} reaches similar performance to our method, it fails to distinguish the ambiguity between albedo and roughness. }}\label{tb:supp_real}
    \scalebox{0.85}{
    \begin{tabular}{*{13}{c}}
      \toprule
      \multirow{2}*{Method} & \multicolumn{3}{c}{InvRender~\cite{zhang2022invrender}} & \multicolumn{3}{c}{NVDIFFREC~\cite{munkberg2021nvdiffrec}} & \multicolumn{3}{c}{NeILF~\cite{yao2022neilf}} & \multicolumn{3}{c}{Ours}\\
      \cmidrule(lr){2-4}\cmidrule(lr){5-7}\cmidrule(lr){8-10}\cmidrule(lr){11-13}
      & PSNR$\uparrow$ & SSIM$\uparrow$ & MSE$\downarrow$ & PSNR$\uparrow$ & SSIM$\uparrow$ & MSE$\downarrow$ & PSNR$\uparrow$ & SSIM$\uparrow$ & MSE$\downarrow$ & PSNR$\uparrow$ & SSIM$\uparrow$ & MSE$\downarrow$ \\
      \midrule
      Scene 1  &23.4773&0.8367&0.0045  &24.6780&0.8776&0.0034  &23.5793&0.8405&0.0044 &\textbf{25.5872}&\textbf{0.8984}&\textbf{0.0028}\\
      Scene 2  &22.3096&0.7603&0.0059  &23.6182&0.8092&0.0043  &22.5556&0.7691&0.0056   &\textbf{24.1521}&\textbf{0.8450}&\textbf{0.0038}\\
      Scene 3  &21.8565&0.7959&0.0065  &22.9661&0.8582&0.0050  &21.8175&0.7994&0.0066 &\textbf{25.3452}&\textbf{0.8820}&\textbf{0.0029}\\
      Scene 4  &21.0931&0.7443&0.0078  &22.3015&0.8150&0.0059  &21.0957&0.7464&0.0078  &\textbf{23.0425}&\textbf{0.8451}&\textbf{0.0050}\\
      Scene 5  &23.0713&0.7764&0.0049  &23.8165&0.8012&0.0042  &23.3284&0.7897&0.0046 &\textbf{24.2985}&\textbf{0.8367}&\textbf{0.0037}\\
      Scene 6  &23.0081&0.7885&0.0050  &25.0760&0.8682&0.0031  &22.7081&0.7860&0.0054  &\textbf{26.1958}&\textbf{0.8943}&\textbf{0.0024}\\
      Scene 7  &20.5928&0.7395&0.0087  &22.0116&0.8149&0.0063  &20.5794&0.7512&0.0088  &\textbf{23.1939}&\textbf{0.8481}&\textbf{0.0048}\\
      Scene 8  &20.8998&0.7083&0.0081  &\textbf{25.8481}&\textbf{0.8816}&\textbf{0.0026}  &20.4024&0.6965&0.0091  &25.3344&0.8542&0.0029\\
      Scene 9  &21.2149&0.7474&0.0076  &24.0453&0.8615&0.0039  &20.7916&0.7331&0.0083  &\textbf{24.3945}&\textbf{0.8732}&\textbf{0.0036}\\
      Scene 11 &22.4695&0.7710&0.0057  &23.1026&0.8015&0.0049  &22.4023&0.7747&0.0058  &\textbf{24.5486}&\textbf{0.8461}&\textbf{0.0035}\\
      \midrule
      Mean     &21.9993&0.7668&0.0065  &23.7464&0.8389&0.0044  &21.9260&0.7687&0.0066  &\textbf{24.6093}&\textbf{0.8622}&\textbf{0.0035}\\
      \bottomrule
    \end{tabular}
    }
    
\end{table*}

\begin{figure*}[t]
    \centering
    \begin{overpic}[scale=0.33]{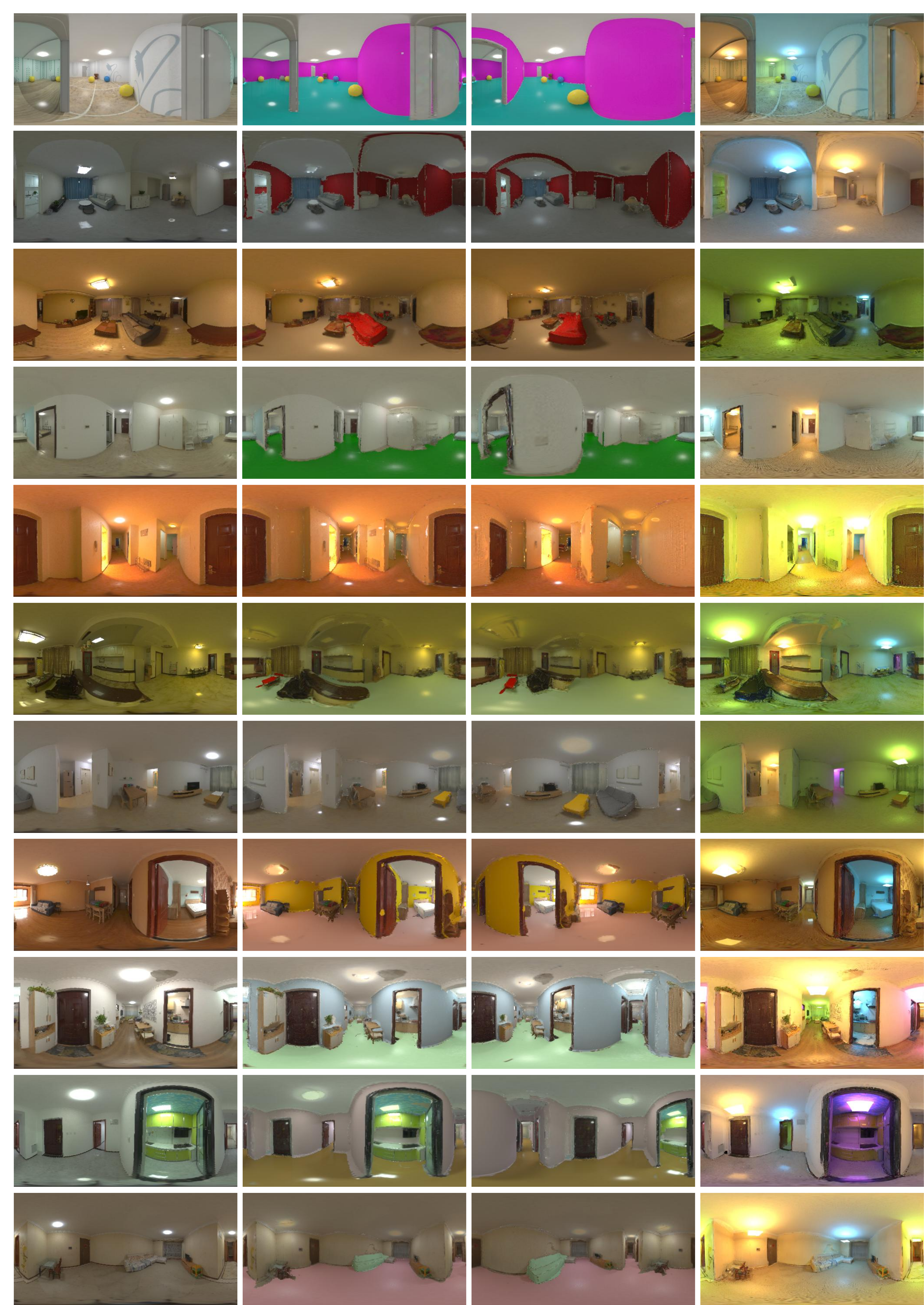}
    
    \put(5, 99.5){\normalsize Source Image}
    \put(22, 99.5){\normalsize Material Editing}
    \put(38, 99.5){\normalsize Editable Novel View}
    \put(58, 99.5){\normalsize Relighting}
    
    \put(-0.3,2){\rotatebox{90} {\footnotesize Scene 11}}
    \put(-0.3,11){\rotatebox{90} {\footnotesize Scene 9}}
    \put(-0.3,20){\rotatebox{90} {\footnotesize Scene 8}}
    \put(-0.3,29){\rotatebox{90} {\footnotesize Scene 7}}
    \put(-0.3,38){\rotatebox{90} {\footnotesize Scene 6}}
    \put(-0.3,47){\rotatebox{90} {\footnotesize Scene 5}}
    \put(-0.3,56){\rotatebox{90} {\footnotesize Scene 4}}
    \put(-0.3,65.5){\rotatebox{90} {\footnotesize Scene 3}}
    \put(-0.3,74){\rotatebox{90} {\footnotesize Scene 2}}
    \put(-0.3,83){\rotatebox{90} {\footnotesize Scene 1}}
    
    \put(-0.3,90.5){\rotatebox{90} {\footnotesize Synthetic Scene}}
    
    \end{overpic}
    \caption{\textbf{Additional samples of applications.} We edit the roughness of floors in Scene 1, Scene 4 and Scene 6, and the albedo for all scenes. Compared to source images, our method still reproduces realistic and consistent lighting effects after editing.}
    \label{fig:supp_app_all}
\end{figure*}
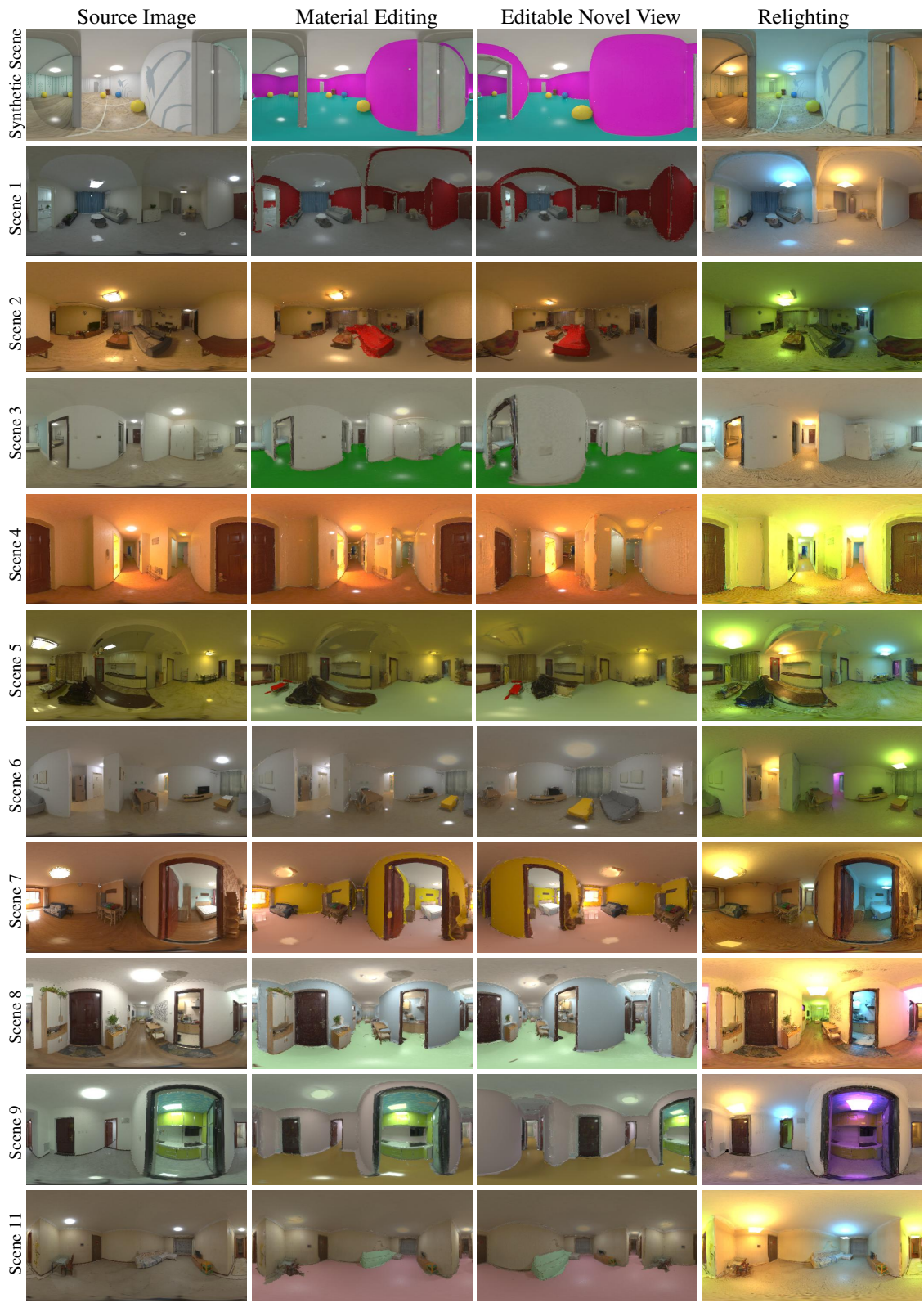

\section{Details of Proposed Datasets}\label{sec:dataset}
\subsection{Synthetic Dataset}

As described in Sec. 4.1 in the main paper, to enable more comprehensive analysis, we create a synthetic scene with diverse material and light sources with a path tracer\cite{li2020inverse}. As shown in Fig.~\ref{fig:syn_data_overview}, the virtual scene is consists of three rooms and several objects with different materials. We generate 40 HDR panoramas, and corresponding poses, semantic segmentation, depth, albedo and roughness annotations, and the entire geometry. 
We use 24 views as input and others as novel views for the novel view synthesis.

\subsection{Full-HDR Real Dataset}
As described in Sec. 4.1 in the main paper, we capture 10 Full-HDR real indoor scenes due to the lack of Full-HDR real dataset. We first use neural networks to predict the corresponding depth images, and leverage SFM and MVS\cite{colmap} to reconstruct the 3D mesh with the RGB texture. As shown in Fig.~\ref{fig:real_data_overview}, 3D indoor scenes are reconstructed. Note that each indoor scene only contains 10 to 20 images. Therefore, the inverse rendering on these real scenes is extreme challenging.

\begin{figure*}[t]
    \centering
    \begin{overpic}[scale=0.25]{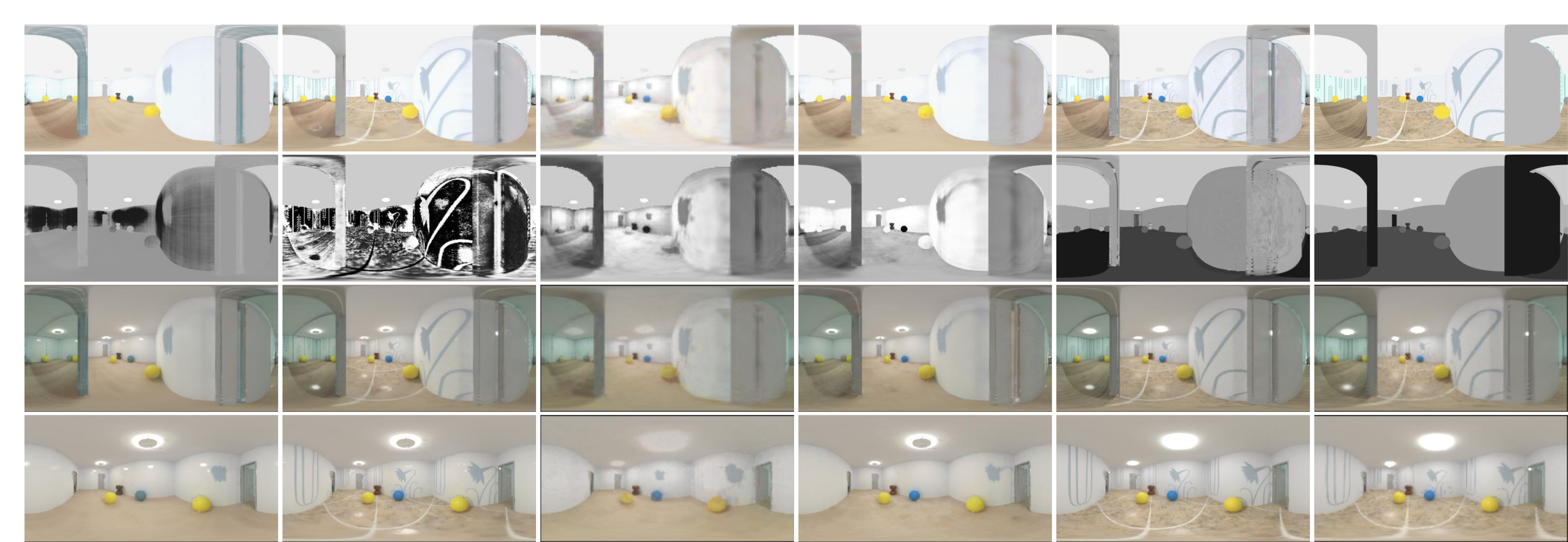}
    \put(4,33.5){\footnotesize InvRender\cite{zhang2022invrender}}
    \put(20,33.5){\footnotesize NVDIFFREC\cite{munkberg2021nvdiffrec}}
    \put(37.5,33.5){\footnotesize NeILF$*$\cite{yao2022neilf}}
    \put(54.5,33.5){\footnotesize NeILF\cite{yao2022neilf}}
    \put(72.5,33.5){\footnotesize Ours}
    \put(89.5,33.5){\footnotesize GT}
    
    \put(0,0){\rotatebox{90} {\footnotesize Novel View}}
    \put(0,9){\rotatebox{90} {\footnotesize Rendering}}
    \put(0,16.6){\rotatebox{90} {\footnotesize Roughness}}
    \put(0,26){\rotatebox{90} {\footnotesize Albedo}}
    
    \end{overpic}
    \caption{\textbf{Additional samples of qualitative comparison on synthetic dataset.} Our method reconstructs globally-consistent and physically-reasonable SVBRDFs while other approaches struggle to reduce ambiguity of materials.}
    \label{fig:comparison_supp_syn}
\end{figure*}

\begin{figure*}[t]
    \centering
    \begin{overpic}[scale=0.25]{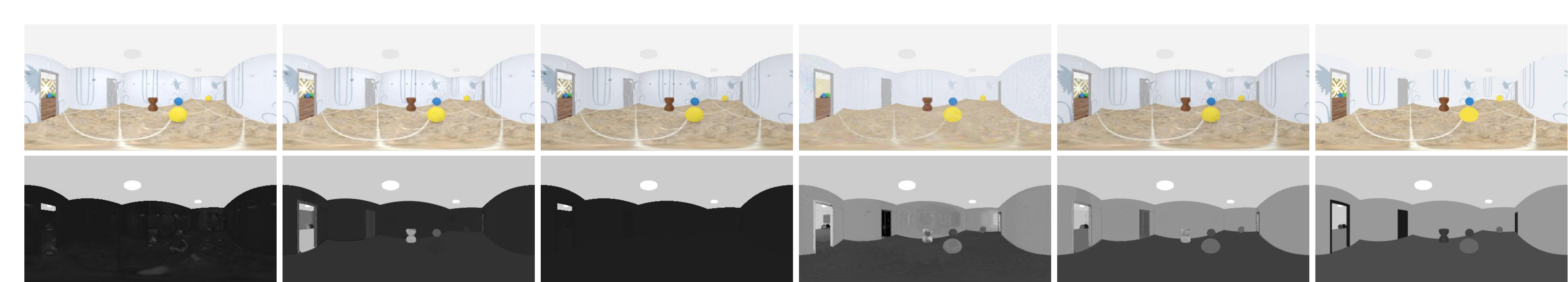}
    \put(6,17){\footnotesize Baseline}
    \put(22,17){\footnotesize w/o Stage \uppercase\expandafter{\romannumeral1} }
    \put(38,17){\footnotesize w/o Stage \uppercase\expandafter{\romannumeral2} }
    \put(55,17){\footnotesize w/o Stage \uppercase\expandafter{\romannumeral3} }
    \put(72.5,17){\footnotesize Ours}
    \put(89.5,17){\footnotesize GT}
    
    \put(0,0.5){\rotatebox{90} {\footnotesize Roughness}}
    \put(0,10){\rotatebox{90} {\footnotesize Albedo}}
    
    \end{overpic}
    \caption{\textbf{Ablation study of three-stage material optimization on synthetic dataset.}}
    \label{fig:ablation_supp_syn}
\end{figure*}

\section{Details of Experiments}\label{sec:experiment}

\subsection{Postprocessing}
We change the albedo and roughness of ceiling and lamps as a postprocessing on synthetic dataset. We empirically found that the predictions of each approach on these regions are easily prone to local minimal. Based on the observation that the roughness and albedo of ceiling is high in most scenes, we set the roughness of ceiling as 0.8 and the albedo as 0.9. Please note that we update the results for each method on synthetic dataset.

\subsection{Results on Costs}
We compare the time cost and memory cost of material optimization to the multi-view inverse rendering methods in Tab.~\ref{tb:supp_cost}. Please note that all methods apply our efficient hybrid lighting representation except for NeILF$*$~\cite{yao2022neilf}. With our hybrid lighting representation, the efficiency of NeILF~\cite{yao2022neilf} is significantly improved. In material optimization, our approach achieves the comparable performance on costs to the previous highly efficient method, NVIDFFREC~\cite{munkberg2021nvdiffrec}. The calculation of IrT with a resolution of 1024$\times$ 1024 takes 10 minutes and costs around 2 GB GPU memory. The optimization process of material takes 20 minutes and also costs around 2 GB GPU memory. Note that the differentiable path tracing-based method~\cite{nimierdavid2021material} takes 12 hours per scene with a significant amounts of GPU memory~\cite{nimierdavid2021material}.

\subsection{Additional Results for Applications}
In Sec.4.5 in the main paper, we demonstrate the capability of our method on several mixed-reality applications, such as material editing, editable novel view synthesis and relighting. We show more results on these applications in Fig.~\ref{fig:supp_app_all}. Benefiting from our triangle mesh and PBR materials output, which is compatible with standard engines, we can easily edit the properties in a physical manner. We change the albedo or roughness according to the semantic segmentation, \eg, the wooden floors become ceramic floors by changing the albedo of floors. Furthermore, we are able to render physically-reasonable novel views based on our 3D geometry and material textures, which is orthogonal to material editing, as shown in third column in Fig.~\ref{fig:supp_app_all}. Last but not least, the entire scene can be rendered under new different illumination, as shown in last column in Fig.~\ref{fig:supp_app_all}. Please refer to supplementary videos for more animations.

\begin{figure}[t]
    \centering
    \begin{overpic}[scale=0.16]{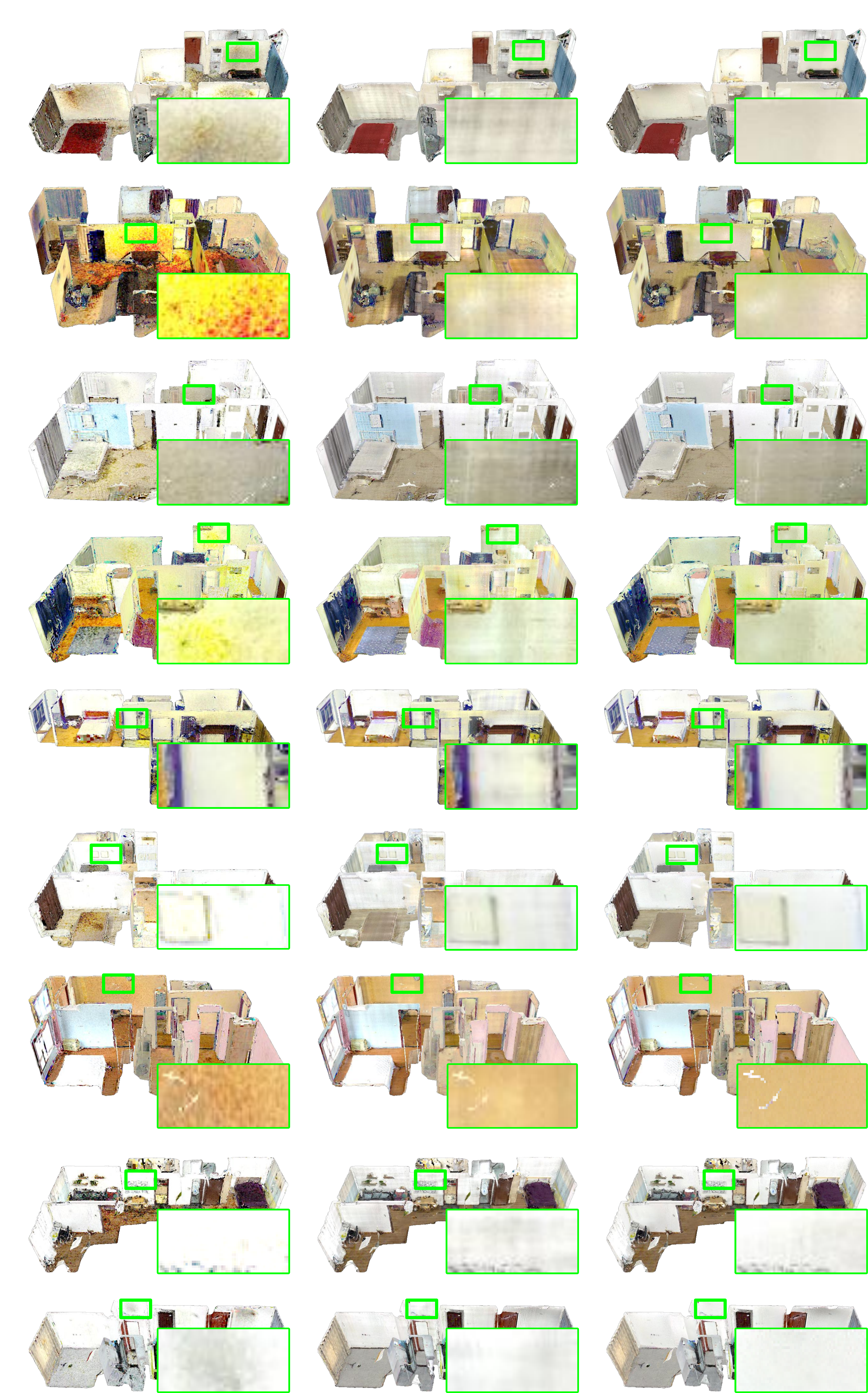}
    
    \put(5, 98){\footnotesize Source TBL}
    \put(30, 98){\footnotesize NIrF}
    \put(51, 98){\footnotesize IrT}

    \end{overpic}
    \caption{\textbf{Ablation study of hybrid lighting representation.} From top to down: Scene 1, Scene 2, Scene 3, Scene 4, Scene 5, Scene 6, Scene 7, Scene 8 and Scene 9. IrT recovers detailed albedo with less artifacts.}
    \label{fig:supp_ablation_irt}
\end{figure}

\subsection{Additional Results on Synthetic Dataset}
We provide more qualitative comparisons on synthetic dataset in Fig.~\ref{fig:comparison_supp_syn}. Our approach is superior than other inverse rendering methods on roughness estimation. And our physically-reasonable and globally-consistent SVBRDFs are able to produce realistic novel views. Note that NeILF~\cite{yao2022neilf} with our hybrid lighting representation more successfully disentangles the ambiguity between materials and lighting than NeILF$*$\cite{yao2022neilf} with their implicit lighting representation.

\begin{figure*}[t]
    \centering
    \begin{overpic}[scale=0.21]{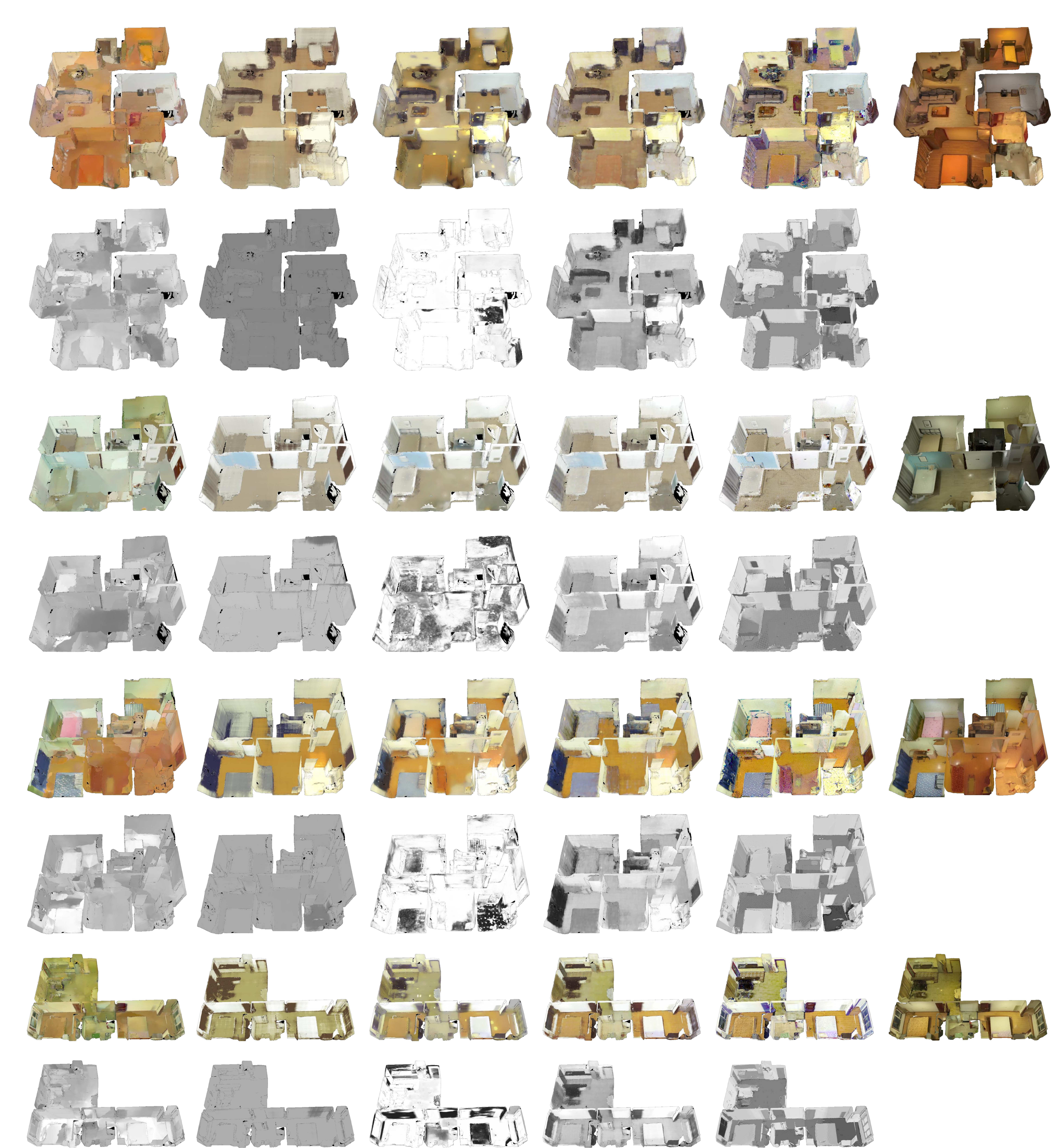}
    \put(6,98){\normalsize PhyIR\cite{li2022phyir}}
    \put(19.5,98){\normalsize InvRender\cite{zhang2022invrender}}
    \put(33,98){\normalsize NVDIFFREC\cite{munkberg2021nvdiffrec}}
    \put(50,98){\normalsize NeILF\cite{yao2022neilf}}
    \put(67,98){\normalsize Ours}
    \put(82,98){\normalsize RGB}
    
    \put(1,0){\rotatebox{90} {\normalsize Roughness}}
    \put(1,10){\rotatebox{90} {\normalsize Albedo}}
    \put(1,20){\rotatebox{90} {\normalsize Roughness}}
    \put(1,32){\rotatebox{90} {\normalsize Albedo}}
    
    \put(1,44){\rotatebox{90} {\normalsize Roughness}}
    \put(1,57){\rotatebox{90} {\normalsize Albedo}}
    \put(1,71){\rotatebox{90} {\normalsize Roughness}}
    \put(1,88){\rotatebox{90} {\normalsize Albedo}}
    
    \end{overpic}
    \caption{\textbf{Additional samples of qualitative comparison in the 3D mesh view on challenging real dataset.} From top to down: Scene 2, Scene 3, Scene 4 and Scene 5. Our method reconstructs globally-consistent and physically-reasonable SVBRDFs while other approaches struggle to produce inconsistent results and reduce ambiguity of materials.}
    \label{fig:comparison_supp_real1}
\end{figure*}

\begin{figure*}[t]
    \centering
    \begin{overpic}[scale=0.21]{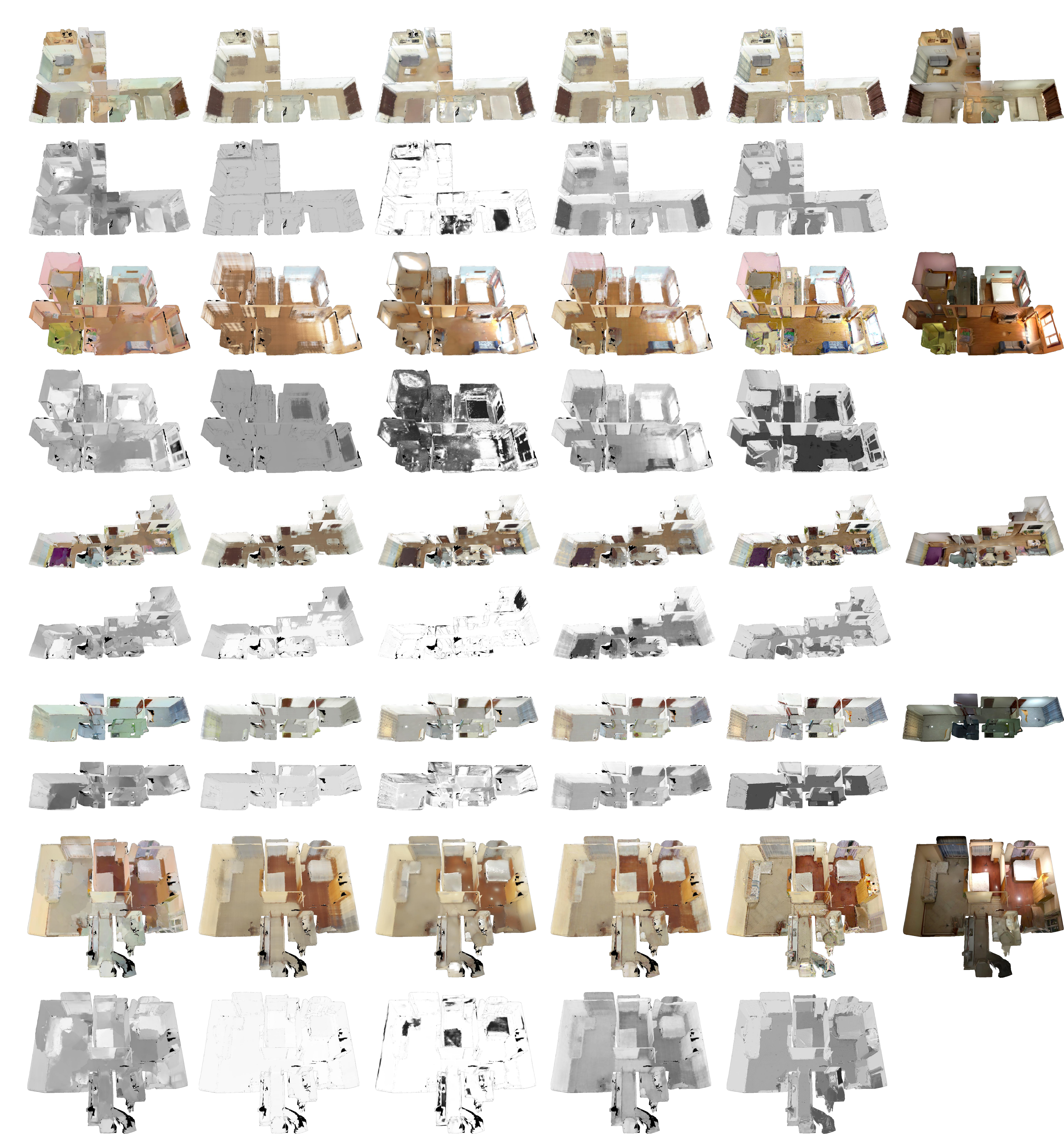}
    \put(5,98){\normalsize PhyIR\cite{li2022phyir}}
    \put(19,98){\normalsize InvRender\cite{zhang2022invrender}}
    \put(34,98){\normalsize NVDIFFREC\cite{munkberg2021nvdiffrec}}
    \put(52,98){\normalsize NeILF\cite{yao2022neilf}}
    \put(69,98){\normalsize Ours}
    \put(85,98){\normalsize RGB}
    
    \put(1,4){\rotatebox{90} {\footnotesize Roughness}}
    \put(1,18){\rotatebox{90} {\footnotesize Albedo}}
    \put(1,27){\rotatebox{90} {\footnotesize Roughness}}
    \put(1,34){\rotatebox{90} {\footnotesize Albedo}}
    
    \put(1,42){\rotatebox{90} {\footnotesize Roughness}}
    \put(1,51){\rotatebox{90} {\footnotesize Albedo}}
    \put(1,59){\rotatebox{90} {\footnotesize Roughness}}
    \put(1,71){\rotatebox{90} {\footnotesize Albedo}}
    \put(1,80){\rotatebox{90} {\footnotesize Roughness}}
    \put(1,91){\rotatebox{90} {\footnotesize Albedo}}
    
    \end{overpic}
    \caption{\textbf{Additional samples of qualitative comparison in the 3D mesh view on challenging real dataset.} From top to down: Scene 7, Scene 8, Scene 9 and Scene 11. Our method reconstructs globally-consistent and physically-reasonable SVBRDFs while other approaches struggle to produce inconsistent results and reduce ambiguity of materials.}
    \label{fig:comparison_supp_real2}
\end{figure*}

\begin{figure*}[t]
    \centering
    \begin{overpic}[scale=0.24]{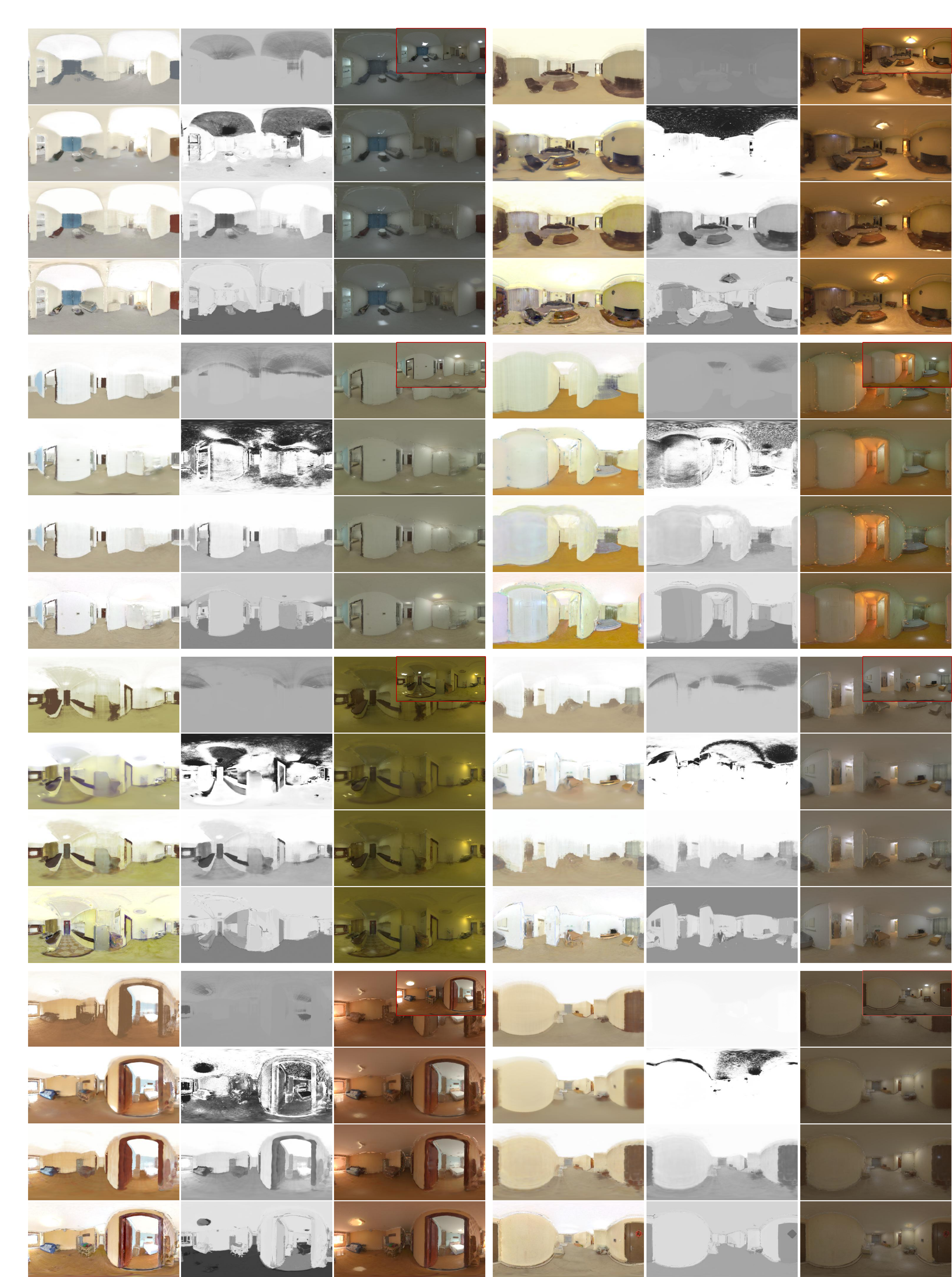}
    
    \put(6, 98){\footnotesize Albedo}
    \put(17, 98){\footnotesize Roughness}
    \put(29, 98){\footnotesize Rendering}
    \put(42, 98){\footnotesize Albedo}
    \put(54, 98){\footnotesize Roughness}
    \put(66, 98){\footnotesize Rendering}
    
    \put(1,2){\rotatebox{90} {\tiny Ours}}
    \put(1,7){\rotatebox{90} {\tiny NeILF\cite{yao2022neilf}}}
    \put(1,12){\rotatebox{90} {\tiny NVDIFFREC\cite{munkberg2021nvdiffrec}}}
    \put(1,18.5){\rotatebox{90} {\tiny InvRender\cite{zhang2022invrender}}}
    
    \put(1,26.5){\rotatebox{90} {\tiny Ours}}
    \put(1,31.5){\rotatebox{90} {\tiny NeILF\cite{yao2022neilf}}}
    \put(1,36){\rotatebox{90} {\tiny NVDIFFREC\cite{munkberg2021nvdiffrec}}}
    \put(1,43){\rotatebox{90} {\tiny InvRender\cite{zhang2022invrender}}}
    
    \put(1,51){\rotatebox{90} {\tiny Ours}}
    \put(1,56){\rotatebox{90} {\tiny NeILF\cite{yao2022neilf}}}
    \put(1,61){\rotatebox{90} {\tiny NVDIFFREC\cite{munkberg2021nvdiffrec}}}
    \put(1,67.5){\rotatebox{90} {\tiny InvRender\cite{zhang2022invrender}}}
    
    \put(1,76){\rotatebox{90} {\tiny Ours}}
    \put(1,80.5){\rotatebox{90} {\tiny NeILF\cite{yao2022neilf}}}
    \put(1,86){\rotatebox{90} {\tiny NVDIFFREC\cite{munkberg2021nvdiffrec}}}
    \put(1,92){\rotatebox{90} {\tiny InvRender\cite{zhang2022invrender}}}
    
    \end{overpic}
    \caption{\textbf{Additional samples of qualitative comparison in the 2D image view on challenging real dataset.} From left to right and from top to down: Scene1, Scene2, Scene3, Scene4, Scene5, Scene6, Scene7 and Scene 11. \textcolor[RGB]{200,0,0}{Red} denotes the Ground Truth image.}
    \label{fig:supp_rendering}
\end{figure*}

\subsection{Additional Results on Real Dataset}
As shown in Tab. 2 in the main paper, our approach outperforms previous neural rendering methods. The detailed results of each real scene are shown in Tab.~\ref{tb:supp_real}. Note that we do not compare to PhyIR~\cite{li2022phyir} with re-rendering error because it uses LDR panoramas as input. Although NVDIFFREC~\cite{munkberg2021nvdiffrec} reaches competitive performance to our method, it fails to distinguish the ambiguity between albedo and roughness in Fig.~\ref{fig:comparison_supp_real1}, Fig.~\ref{fig:comparison_supp_real2} and Fig.~\ref{fig:supp_rendering}. Our approach is able to reconstruct physically-reasonable and globally-consistent SVBRDF. Such properties re-render similar specular reflectance to GT with less wrong highlights in albedo, which proves we disentangle the ambiguity of materials successfully.

\begin{figure*}[t]
    \centering
    \begin{overpic}[scale=0.21]{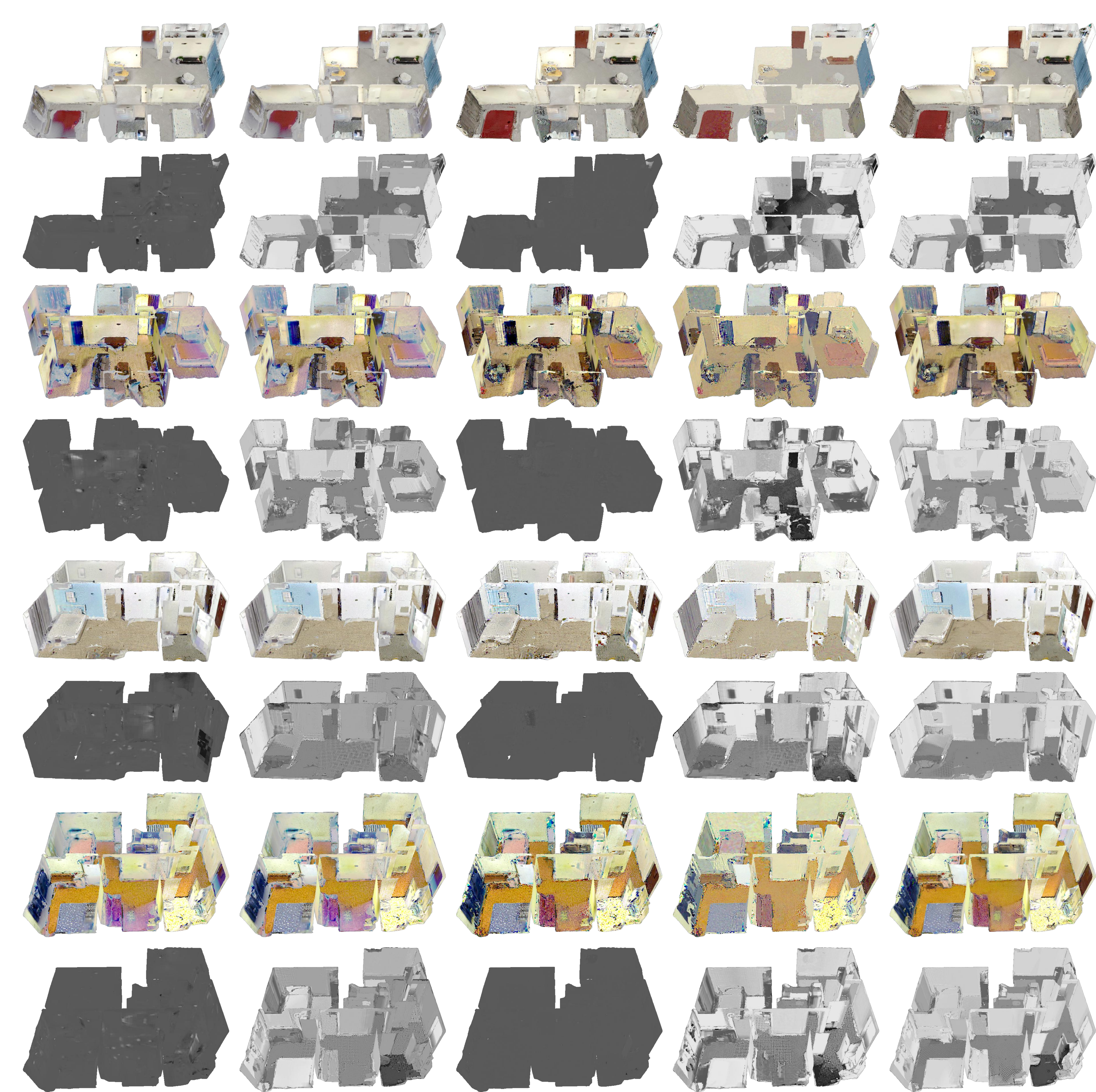}
    
    \put(8, 98){\normalsize Baseline}
    \put(27, 98){\normalsize w/o Stage \uppercase\expandafter{\romannumeral1} }
    \put(47, 98){\normalsize w/o Stage \uppercase\expandafter{\romannumeral2} }
    \put(66, 98){\normalsize w/o Stage \uppercase\expandafter{\romannumeral3} }
    \put(89, 98){\normalsize Ours}
    
    \put(0,3){\rotatebox{90} {\normalsize Roughness}}
    \put(0,19){\rotatebox{90} {\normalsize Albedo}}
    \put(0,29){\rotatebox{90} {\normalsize Roughness}}
    \put(0,42){\rotatebox{90} {\normalsize Albedo}}
    
    \put(0,52){\rotatebox{90} {\normalsize Roughness}}
    \put(0,66){\rotatebox{90} {\normalsize Albedo}}
    \put(0,76){\rotatebox{90} {\normalsize Roughness}}
    \put(0,89){\rotatebox{90} {\normalsize Albedo}}
    
    \end{overpic}
    \caption{\textbf{Additional samples of ablation study of material optimization on challenging real dataset.} From top to down: Scene 1, Scene 2, Scene 3 and Scene 4.}
    \label{fig:supp_ablation_material_1}
\end{figure*}

\begin{figure*}[t]
    \centering
    \begin{overpic}[scale=0.21]{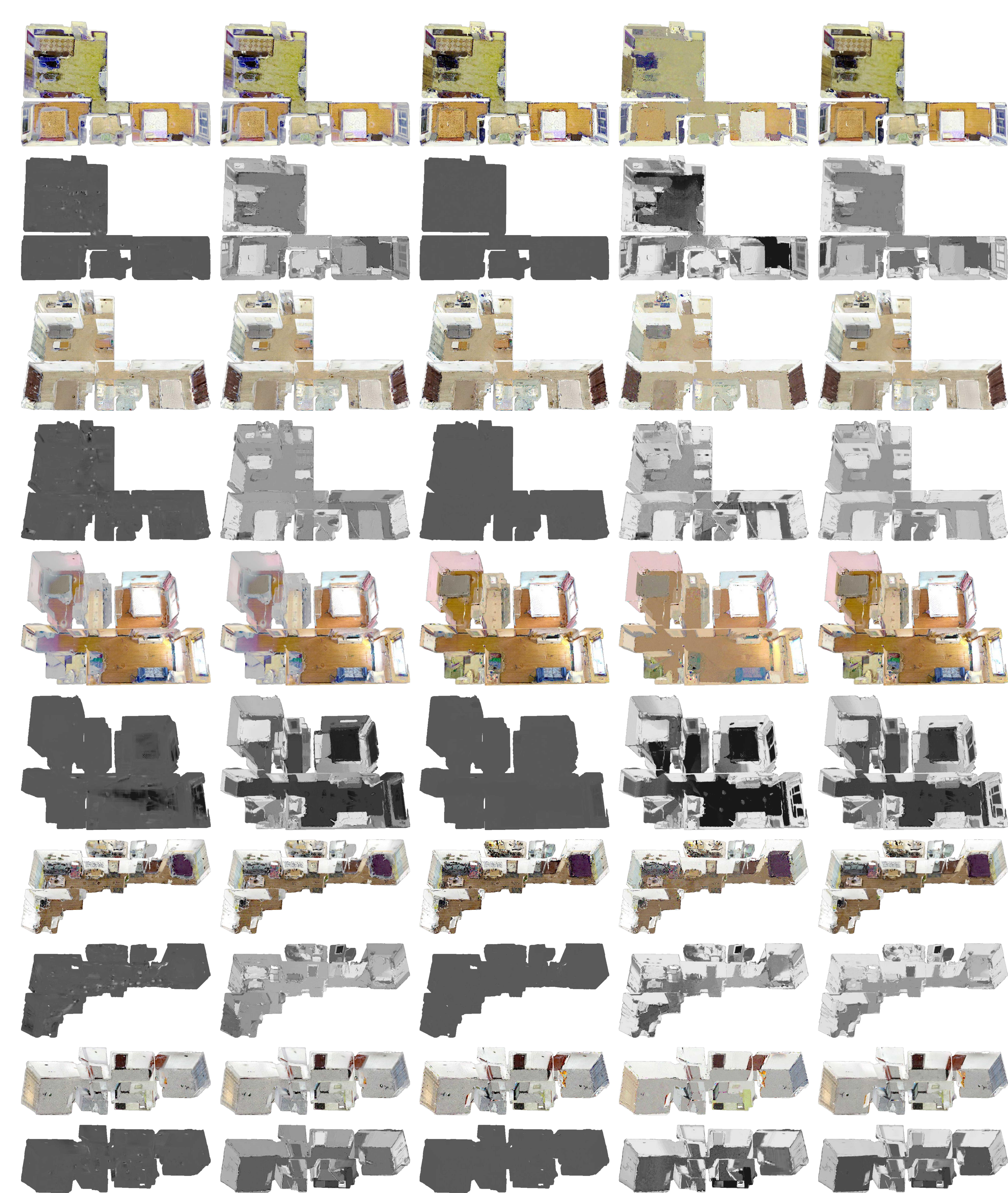}
    
    \put(7, 99){\normalsize Baseline}
    \put(23, 99){\normalsize w/o Stage \uppercase\expandafter{\romannumeral1} }
    \put(39, 99){\normalsize w/o Stage \uppercase\expandafter{\romannumeral2} }
    \put(56, 99){\normalsize w/o Stage \uppercase\expandafter{\romannumeral3} }
    \put(75, 99){\normalsize Ours}
    
    \put(0,-1){\rotatebox{90} {\normalsize Roughness}}
    \put(0,7){\rotatebox{90} {\normalsize Albedo}}
    \put(0,14){\rotatebox{90} {\normalsize Roughness}}
    \put(0,24){\rotatebox{90} {\normalsize Albedo}}
    
    \put(0,33){\rotatebox{90} {\normalsize Roughness}}
    \put(0,46){\rotatebox{90} {\normalsize Albedo}}
    \put(0,56){\rotatebox{90} {\normalsize Roughness}}
    \put(0,69){\rotatebox{90} {\normalsize Albedo}}
    \put(0,78){\rotatebox{90} {\normalsize Roughness}}
    \put(0,91){\rotatebox{90} {\normalsize Albedo}}
    
    \end{overpic}
    \caption{\textbf{Additional samples of ablation study of material optimization on challenging real dataset.} From top to down: Scene 1, Scene 2, Scene 3 and Scene 4.}
    \label{fig:supp_ablation_material_2}
\end{figure*}

\subsection{Additional Results for Ablation studies}
We showcase the effectiveness of our three-stage material optimization on synthetic dataset in Fig.~\ref{fig:ablation_supp_syn}. As described in Sec. 4.4 in the main paper, the Baseline only update the highlight regions of roughness. Without Stage \uppercase\expandafter{\romannumeral1}, the roughness leads to incorrect result. Without Stage \uppercase\expandafter{\romannumeral2}, the performance of roughness estimation will decrease dramatically. Without Stage \uppercase\expandafter{\romannumeral3}, the abledo is over-blur and the roughness is unsmooth.

As shown in Fig. 7 in the main paper, we show one sample for ablating the effectiveness of hybrid lighting representation. We show more results in Fig.~\ref{fig:supp_ablation_irt}. The proposed IrT recovers detailed albedo with less noise.

Additionally, we show more ablation studies of our material optimization strategy on real dataset in Fig.~\ref{fig:supp_ablation_material_1} and Fig.~\ref{fig:supp_ablation_material_2}. 

Finally, we show the performance of our method as the semantic segmentation mask becomes less accurate. We randomly change a cube region with wrong semantic labels for each input image. As shown in Tab.~\ref{tb:ablation_syn}, our method is surprisingly robust as the length of cube increases.

\begin{table}[t]
	\centering \caption{\textbf{Ablation study of the quality of semantics.} }\label{tb:ablation_syn}
    \scalebox{0.6}{
    \begin{tabular}{ccccccc}
       \toprule
       Property (PSNR) & 0*0 & 16*16 & 32*32 & 64*64 & 128*128 & 256*256 \\
       \midrule
       Albedo & 20.4169 & 20.7858  & 20.7353 &21.0199 &20.8364  &19.7991\\
       Roughness & 20.2132 & 19.8076 & 19.9088 &19.8964 &17.8038 &13.5650 \\
       \bottomrule
    \end{tabular}
    }
\end{table}

\subsection{Bad Cases}
As described in Sec. 4.6 in the main paper, our method lead to recover bright albedo and low roughness when the light source is not captured. In Scene 8 in the Fig.~\ref{fig:comparison_supp_real2}, we reconstruct over-high albedo and over-low roughness nearby the window because the sun is not captured. The learning prior will be helpful for disentangling the ambiguity between materials in such cases.

\section{More Discussions}
\label{sec:discussion}
\subsection{Limitations and Future works}
There are some limitations of our method. First, we rely on the HDR images to recover the proposed lighting representation for large-scale scene. To lift this limitation, the joint optimization of lighting and material will be explored.
Second, our VHL-based sampling and semantics-based propagation requires that light sources are visible in the scene. If light sources are not captured, our method leads to recover bright albedo and low roughness. In such cases, we have to leverage the learning prior to alleviate the ambiguity of materials.
Finally, although the geometry reconstructed by MVS is enough for our method, a more accurate geometry would lead to more accurate predictions.

\subsection{TBL and Path tracing}
The main pros of TBL is much less time and memory costs, compared to the path tracer~\cite{azinovic2019inverse, nimierdavid2021material}. Our method only takes 30 minutes while~\cite{nimierdavid2021material} takes 12 hours per scene, reported in their paper. Moreover, the accuracy and robustness of TBL also is higher than the path tracer. If the recursive rendering equation can be computed instantly, the high gradient caused by the recursion and low samples in path sampling still do not ensure steady convergence~\cite{nimierdavid2021material}. On the one hand, our TBL models the complex light transport as a relatively simple local shading, which ensures more robust optimization. On the other hand, the global illumination of path tracing is finite-bounce while the TBL represents infinite-bounce global illumination, corresponding to real world. Therefore, the global illumination of TBL is more accurate.

In some cases, both our TBL and the path tracer do not work well, \eg, some important light sources or regions are missing, transparent/translucent objects, participating media and caustics. The differentiable volume rendering and neural rendering will be nice choices for such hard cases. I agree that some effects, \eg, a chain of specular reflections and retroreflections could be solved well using a path tracer while our TBL fails to model such effects. However, such effects are rare in most indoor scenes.

\subsection{Broader Impacts}
As described in the main paper, our method is able to produce realistic and physically-reasonable images with modified materials or illumination. Therefore, creating deepfake is a major potential negative impact. We can limit the target scenarios to prevent malicious use cases.

\end{document}